\documentclass{tlp}
\usepackage{aopmath}

\def\extendedPaper{\def\extPaper{T}}
\def\extPaper{}

\extendedPaper

\usepackage{epsfig}

\def\no{{not} \;} 
\def\beq{\begin{equation}} 
\def\eeq#1{\label{#1}\end{equation}} 
\def\st{\smallskip\noindent}

\newcommand{\AL}{${\mathcal A}{\mathcal L}$ }
\newcommand{\ALns}{${\mathcal A}{\mathcal L}$}
\newcommand{\tbeg}{\langle} 
\newcommand{\tend}{\rangle} 
\newcommand{\entails}{\models}

\newcommand{\lpnot}{\mbox{not}\;\,}

\newcommand{\hif}{\leftarrow}

\newcommand{\eqO}[1]{\simeq_{#1}}

\renewcommand{\land}{\mbox{ and }}

\newcommand{\Begproof}{\begin{proof}}
\newcommand{\Endproof}{\end{proof}}

\newtheorem{example}{Example}[section]
\newtheorem{definition}{Definition}[section]

\title[Diagnostic reasoning with A-Prolog]{Diagnostic reasoning with A-Prolog {\normalsize\thanks{
This work was supported in part by United Space Alliance under Research Grant 26-3502-21 and Contract COC671311, and
by NASA under Contracts 1314-44-1476 and 1314-44-1769. An extended version of this paper is available from
\texttt{http://www.krlab.cs.ttu.edu}. }}}

\author[M. Balduccini and M. Gelfond]
{MARCELLO BALDUCCINI and MICHAEL GELFOND\\
Department of Computer Science, Texas Tech University\\
Lubbock, TX  79409, USA\\ 
\email{\{mgelfond,balduccini\}@cs.ttu.edu}
}

\begin{document}
\maketitle

\begin{abstract}
In this paper we suggest  
an architecture for a software agent which
operates a physical device and is
capable of making observations and of testing and repairing
the device's components. We present simplified definitions of the
notions of symptom, candidate diagnosis, and diagnosis which
are based on the theory of action language ${\cal AL}$.
The definitions allow one to give a simple account of
the agent's behavior in which many of the agent's tasks 
are reduced to computing stable models of logic programs.
\end{abstract}


\section{Introduction}
In this paper we continue the investigation
of applicability of A-Prolog (a loosely defined
collection of logic programming languages under the answer 
set (stable model) semantics \cite{gl88,gl91} to knowledge 
representation and reasoning.
The focus is on the development of an architecture for a software 
agent acting in a changing environment.
We assume that the agent and the environment (sometimes
referred to as a dynamic system) satisfy the following simplifying
conditions.
\begin{enumerate}
\item The agent's environment can be viewed as a transition
diagram whose states are sets of fluents 
(relevant properties of the domain whose truth values may depend on time)
 and whose arcs are labeled by actions.
\item The agent is capable of making correct observations, performing
actions, and remembering the domain history.
\item Normally the agent is capable of observing all relevant exogenous
events occurring in its environment.
\end{enumerate}
These assumptions hold in many realistic domains and are suitable
for a broad class of applications. In many domains, however, the
effects of actions and the truth values of observations can only be known
with a substantial  degree of uncertainty which cannot be ignored
in the modeling process. It remains to be seen if some of our 
methods can be made to work in such situations.
The above assumptions  determine the structure of
the agent's knowledge base. It consists of three parts. The {\em
first part},  called an {\em action} (or {\em system})
{\em description},
 specifies the
transition diagram representing possible trajectories of the system. 
It contains descriptions of domain's
actions and fluents, together with the
definition of possible successor states to which the system can
move after an action $a$ is executed in a state $\sigma$. 
The {\em
second part} of the agent's knowledge, called a {\em recorded history}
contains observations made
by the agent together with a record of its own actions. It defines
a collection of paths in the diagram which, from the standpoint of the
agent, can be interpreted as
the system's possible pasts. If the
agent's knowledge is complete (e.g., it has complete information
about the initial state and the occurrences of actions, and the system's
actions are deterministic) then there is only one such path. 
The {\em third part} of agent's knowledge base
contains a collection of the agent's goals. All  
this knowledge is used and updated by the agent who  
repeatedly executes the following steps (the \emph{observe-think-act-loop} \cite{ks99,bg00}):
\begin{enumerate}
\item observe the world and interpret the observations;
\item  select a goal;
\item  plan;
\item  execute part of the plan.
\end{enumerate}
In this paper we concentrate on agents operating
physical devices and capable of testing and repairing
the device components. We are especially interested in 
the first step of the loop, i.e. in agent's interpretations 
of discrepancies between agent's expectations and the system's actual behavior.
The following example will be used throughout the paper:
\begin{figure}[tb]
\begin{center}
\includegraphics{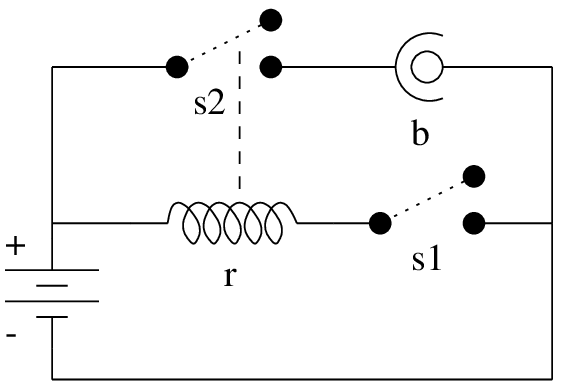}  
\caption{$\cal{AC}$}\label{fig.output}
\end{center}
\end{figure}
\begin{example}\label{ex1}
{\rm Consider a system $S$ consisting of an agent operating an
analog circuit $\cal {AC}$ from figure 1.
We assume that switches $s_1$ and $s_2$ are mechanical components which 
cannot become damaged.  Relay {\em r} is a magnetic coil. 
If not damaged, it is activated when $s_1$ is closed, 
causing $s_2$ to  close.  Undamaged
bulb {\em b} emits light if $s_2$ is closed. 
For simplicity of presentation
we consider the agent capable of performing only one action,
$close(s_1)$. The environment can be represented
by two damaging exogenous\footnote{By {\em exogenous} actions
we mean actions performed by the agent's environment. This includes natural 
events as well as actions performed by other agents.}
actions:
$brk$, which causes {\em b} to become faulty, and 
$srg$ (power surge), which damages {\em r} and also {\em b} 
assuming that {\em b} is not protected. 
Suppose that the agent operating this device is given a goal
of lighting the bulb. He realizes that this can be achieved
by closing the first switch, performs the operation,
and discovers that the bulb is not lit. The {\em goal of the paper 
is to develop methods for modeling the agent's behavior after this discovery}.
}
\end{example}

\st
We start with presenting a mathematical model of an agent
and its environment based on the theory of action languages \cite{gl98}.
Even though our approach is applicable to a large collection of action
languages, to simplify the discussion we
will limit our attention to action  language 
${\cal AL}$ from \cite{bg00}. 
We proceed by presenting definitions of the
notions of symptom, candidate diagnosis, and diagnosis which
somewhat differ from those we were able to find in the literature.
These definitions are used to give a simple account of
the agent's behavior including diagnostics, testing, and repair.
We also suggest algorithms for performing these tasks, which are
based on encoding the agents knowledge in A-Prolog
and reducing the agent's tasks to computing stable models
(answer sets) of logic programs.

In this paper we assume that at any moment of time
the agent is capable of testing whether a given component is functioning
properly. Modification of the algorithms in the situation when this
assumption is lifted is the subject of further research.

\st
There is a numerous literature on automating various
types of diagnostic tasks and the authors were greatly influenced by it. 
We mention only several papers which served as a starting point
for our investigation. Of course we are indebted to R. Reiter \cite{r87} 
which seems to contain the first clear logical account of the diagnostic
problem. We were also influenced by early papers of D. Poole and K. Eshghi
who related diagnostics and logic programming, seriously discussed
the relationship between diagnostics and knowledge representation,
and thought about the ways to combine descriptions of normal behaviour
of the system with information about its faults.
More recently M. Thielscher, S. McIlraith, 
C. Baral, T. Son, R. Otero 
recognized that diagnostic problem solving 
involves reasoning about the evolution of dynamic systems,
related diagnostic reasoning with reasoning about action, change, and
causation, and told the story of diagnostics which included 
testing and repair.

\st
In our paper we generalize and modify this work in several directions. 

\begin{itemize} 
\item We considered a simple and powerful language ${\cal AL}$
for describing the agent's knowledge. Unlike some of the previous languages
used for this purpose, ${\cal AL}$ allows concurrent actions and 
consecutive time-steps, and makes the distinction between observations
and the derived (possibly defeasible) knowledge. 
The semantics of the language allows to explain malfunctioning of the system
by some past occurrences of exogenous (normally breaking) actions which
remain unobserved by the agent. 

\item We simplified the basic definitions such as 
symptom, candidate diagnosis, and diagnosis. 

\item We established the realtionship between ${\cal AL}$ and logic programming
and used this relationship to reduce various diagnostic tasks 
to computing stable models of logic programs. 

\item Finally we proved correctness of the corresponding diagnostic algorithms.
\end{itemize}

\st
The paper is organized as follows: in Section \ref{sec:modeling} we introduce a
motivating example. Section \ref{sec:basic-defs} introduces basic definitions used throughout
the paper. In Sections \ref{sec:computing} and \ref{sec:finding}, we show how
techniques of answer set programming can be applied to the computation of candidate diagnoses
and of diagnoses. In Section \ref{sec:repair} we investigate the issues related to the introduction
of the ability to repair damaged components. Section \ref{sec:rel-work} discusses
related work. In Section \ref{sec:conclusions} we conclude the paper and
describe how our work can be extended. The remaining sections contain the description of
syntax and semantics of A-Prolog and \ALns, as well as the proofs of the main theorems stated in
this paper.

\section{Modeling the domain}\label{sec:modeling}
We start with some formal definitions describing a diagnostic domain 
consisting of an agent controlling a physical device.
We limit ourselves to {\em non-intrusive} and {\em observable}
domains in which the {\em agent's environment does not
normally interfere with his work} and {\em the agent
normally observes all of the domain occurrences of exogenous actions}.
The agent is, however, aware of the fact that
these assumptions can be contradicted by
observations. As a result the agent is ready to observe and to take into
account occasional occurrences of exogenous `breaking' actions.
Moreover, discrepancies between expectations and observations
may force him to conclude that some
exogenous actions in the past remained unobserved. 
This view of the relationship between the agent and his environment
determined our choice of action language used
for describing the agent's domain and, to the large extent, is responsible
for substantial differences between our approach and that of \cite{bms00}.

\st
By a {\em domain signature} we mean a triple $\Sigma=\langle C,F,A \rangle$ 
of disjoint finite sets. Elements of $C$ will be
called device {\em components} and used to name various parts of the
device. Elements of $F$ are referred to as {\em fluents} and used
to denote dynamic properties of the domain 
\footnote{Our definitions could be easily
generalized to domains with non-boolean fluents. However, the restriction
to boolean fluents will simplify the
presentation.}.
By {\em fluent literals} we mean fluents 
and their negations (denoted  by $\neg f$).
We also assume existence of a set $F_0 \subseteq F$ which, 
intuitively, corresponds to the class of fluents which can be directly
observed by the agent. 
The set of literals formed from a set $X \subseteq F$ of fluents will be
denoted by $lit(X)$. A set $Y \subseteq lit(F)$ is called {\em complete}
if for any $f \in F$, $f \in Y$ or $\neg f \in Y$; $Y$ is called {\em
  consistent} if there is no $f$ such that $f, \neg f \in Y$.
We assume that for every component $c$ the set $F_0$ contains
a fluent $ab(c)$ which says that the device's component $c$ 
is faulty. The use of $ab$ in diagnosis goes back to \cite{r87}.
The set $A$ of {\em elementary actions} is partitioned into two 
disjoint sets, $A_s$ and $A_e$; $A_s$ consists of {\em actions performed 
by an agent} and $A_e$ consists of {\em exogenous actions}. 
(Occurrences of unobserved exogenous actions will be viewed as 
possible causes of the system's malfunctioning).

\st
By a {\em transition diagram} over signature $\Sigma$ we mean
a directed graph $T$ such that:

\st
(a) the states of $T$ are labeled by complete and consistent 
sets of fluent literals (corresponding to possible physical states of 
the domain).

\st
(b) the arcs of $T$ are labeled by subsets of $A$ called {\em compound actions}.
(Intuitively, execution of a compound action $\{a_1,\dots,a_k\}$ corresponds to 
the simultaneous execution of its components).

\st
Paths of a transition diagram correspond to {\em possible trajectories}
of the domain. A particular trajectory, $W$, called the {\em actual trajectory}
corresponds to the actual behavior of the domain. 
In our observe-think-act loop the agent's connection with reality is 
modeled by a function $observe(n,f)$ which
takes a natural number $n$ and a fluent $f \in F_0$ 
as parameters and returns 
$f$ if $f$ belongs to the $n$'th state of $W$ and $\neg f$ otherwise
\begin{definition}\label{dd}
{\rm By a {\em diagnostic domain} we mean a triple 
$\langle \Sigma, T, W \rangle$ where $\Sigma$ is a domain signature, 
$T$ is a transition diagram
over $\Sigma$, and $W$ is the domain's actual trajectory.
}
\end{definition}
To design an intelligent agent associated with a diagnostic domain 
$S =\langle \Sigma,T,W\rangle$ 
we need to supply the agent with the knowledge of $\Sigma$, $T$, 
and the recorded history of $S$ up to a current point $n$.
Elements of $\Sigma$ can normally be defined by a simple logic program.
Finding a concise and convenient way to define the transition diagram
of the domain is somewhat more difficult. We start with limiting our
attention to transition diagrams defined by action descriptions of action
language ${\cal AL}$ from \cite{bg00}. The accurate description
of the language can be found in Section \ref{app2}.
A typical action description $SD$ of ${\cal AL}$ consists of
a collection of {\em causal laws} determining the effects of the domain's
actions, the actions' {\em executability conditions}, and the {\em state
 constraints} - statements describing dependences between fluents.
(We often refer to statements of $SD$ as {\em laws}.)
Causal laws of $SD$ can be divided 
into two parts. The first part, $SD_n$, contains laws describing normal
behavior of the system. Their bodies usually contain special
fluent literals of the form $\neg ab(c)$.
The second part, $SD_b$, describes effects of
exogenous actions damaging the components. Such laws normally
contain relation $ab$ in the head or
positive parts of the bodies.
(To simplify our further discussion we only consider exogenous actions
capable of causing malfunctioning of the system's components.
The restriction is however inessential and can easily be lifted.)

\st
By the {\em recorded history} $\Gamma_n$ 
 of $S$  up to a current moment $n$ we mean a collection of {\em observations},
i.e. statements of the form: 
\begin{enumerate}
\item $obs(l,t)$  - `fluent literal
$l$ was observed to be true at moment $t$';
\item
$hpd(a,t)$ - elementary action $a \in A$ was observed to happen at moment $t$ 
\end{enumerate}
where $t$ is an integer from the interval $[0,n)$.
Notice that, intuitively, recorded history $hpd(a_1,1), hpd(a_2,1)$ says
that an 'empty' action, $\{\}$, occurred at moment 0 and actions $a_1$ and $a_2$
occur concurrently at moment 1.

\st
An agent's knowledge about the domain up to moment $n$ will consists
of an action description of ${\cal AL}$ and domain's recorded history.
The resulting theory will often be referred to as a {\em domain
  description} of ${\cal AL}$.
\begin{definition}\label{model}
{\rm Let $S$ be a diagnostic domain with transition diagram $T$ 
and actual trajectory 
$W = \langle\sigma^{w}_0, a^{w}_0, \sigma^{w}_1, \ldots, 
a^{w}_{n-1}, \sigma^{w}_n\rangle$, and let $\Gamma_n$ be a recorded 
history of $S$ up to moment $n$.

\noindent
(a) A path $\langle \sigma_0, a_0, \sigma_1, \ldots, a_{n-1}, \sigma_n\rangle$
in $T$ is a {\em model} of $\Gamma_n$ (with respect to $S$) if
for any $0 \leq t \leq n$ 
\begin{enumerate}
\item $a_t = \{a : hpd(a,t) \in \Gamma_n\}$;
\item  if $obs(l,t) \in \Gamma_n$ then $l \in \sigma_t$.
\end{enumerate} 

\st
(b) $\Gamma_n$ is {\em consistent} (with respect to $S$) if it has a model.

\st
(c) $\Gamma_n$ is {\em sound} (with respect to $S$) if, for any 
$l$, $a$, and $t$, if $obs(l,t), hpd(a,t) \in \Gamma_n$ then
$l \in \sigma^{w}_t$ and $a \in a^{w}_t$.
 
\st
(d) A fluent literal $l$ {\em holds} in a model $M$ of $\Gamma_n$ at time 
$t \leq n $ 
($M \models h(l,t)$) if $l \in \sigma_t$; $\Gamma_n$ {\em entails} $h(l,t)$ 
($\Gamma_n \models h(l,t)$) if, for every model $M$ of $\Gamma_n$, 
$M \models h(l,t)$.
}
\end{definition}
Notice that, in contrast to definitions from
\cite{bms00} based on action description
language ${\cal L}$ from \cite{bgp94}, recorded history 
in ${\cal AL}$ is consistent only if
changes in the observations of  system's states can be explained
without assuming occurrences of any action not recorded
in $\Gamma_n$. Notice also that a recorded history may be
consistent, i.e. compatible with $T$, but not sound, i.e. incompatible
with the actual trajectory of the domain.

\st
The following is a description, $SD$, of system $S$ from Example
\ref{ex1}:

\[
\hspace*{-0.4in}\begin{array}{rlrl}
\begin{array}{r}
Objects
\end{array} & \hspace*{-0.2in}\left\{
\begin{array}{l}
comp(r)\mbox{.}\\
comp(b)\mbox{.}\\
switch(s_1)\mbox{.}\\
switch(s_2)\mbox{.}
\end{array}
\right. &
\begin{array}{r}
Fluents
\end{array} & \hspace*{-0.2in}\left\{
\begin{array}{l}
fluent(active(r))\mbox{.}\\
fluent(on(b))\mbox{.}\\
fluent(prot(b))\mbox{.}\\
fluent(closed(SW)) \leftarrow switch(SW)\mbox{.}\\
fluent(ab(X)) \leftarrow comp(X)\mbox{.}
\end{array}
\right.\\
\begin{array}{r}
Agent \\
Actions
\end{array} & \hspace*{-0.2in}\left\{
\begin{array}{l}
\\
a\_act(close(s_1))\mbox{.}\\
\\
\end{array}
\right. &
\begin{array}{r}
Exogenous \\
Actions
\end{array} & \hspace*{-0.2in}\left\{
\begin{array}{l}
\\
x\_act(brk)\mbox{.}\\
x\_act(srg)\mbox{.}\\
\\
\end{array}
\right.
\end{array}
\]
%

\st
Causal Laws and Executability Conditions describing normal
functioning of $S$:

\st
$SD_n \: \left\{
\begin{array}{l}
causes(close(s_1),closed(s_1),[])\mbox{.} \\
caused(active(r),[closed(s_1), \neg ab(r)])\mbox{.}\\
caused(closed(s_2), [active(r)])\mbox{.}\\
caused(on(b),[closed(s_2), \neg ab(b)])\mbox{.}\\
caused(\neg on(b),[\neg closed(s_2)])\mbox{.}\\
impossible\_if(close(s_1), [closed(s_1)])\mbox{.}
\end{array}
\right.
$

\st
($causes(A,L,P)$ says that execution of elementary action $A$ in a state
satisfying fluent literals from $P$ causes fluent literal $L$ to become
true in a resulting state; $caused(L,P)$ means that every state satisfying
$P$ must also satisfy $L$, $\\impossible\_if(A,P)$ indicates that action $A$
is not executable in states satisfying $P$.)  
The system's malfunctioning information from Example \ref{ex1} is given by:

\[
SD_b \: \left\{
\begin{array}{ll}
\begin{array}{l}
causes(brk,ab(b),[])\mbox{.}\\
causes(srg, ab(r),[])\mbox{.}\\
causes(srg, ab(b),[\neg prot(b)])\mbox{.}
\end{array} &
\begin{array}{l}
caused(\neg on(b),[ab(b)])\mbox{.}\\
caused(\neg active(r), [ab(r)])\mbox{.}
\end{array}
\end{array}
\right.
\]

\st
Now consider a history, $\Gamma_1$ of $S$:

\[
\Gamma_1 \: \left\{
\begin{array}{ll}
\begin{array}{l}
hpd(close(s_1),0)\mbox{.}\\
obs(\neg closed(s_1),0)\mbox{.}\\
obs(\neg closed(s_2),0)\mbox{.}
\end{array} &
\begin{array}{l}
obs(\neg ab(b),0)\mbox{.}\\
obs(\neg ab(r),0)\mbox{.}\\ 
obs(prot(b),0)\mbox{.}
\end{array}
\end{array}
\right.
\]
$\Gamma_1$ says that, initially, the agent observed that $s_1$ and $s_2$ were open, both
the bulb, $b$, and the relay, $r$, were not to be damaged, and the bulb was protected from surges.
$\Gamma_1$ also contains the observation that action $close(s_1)$ occurred at time $0$.

\st
Let $\sigma_0$ be the initial state, and $\sigma_1$ be the successor state, reached by
performing action $close(s_1)$ in state $\sigma_0$. It is easy to see that the path 
$\langle \sigma_0, close(s_1), \sigma_1 \rangle$ is 
the only model of $\Gamma_1$  and  that $\Gamma_1 \models h(on(b),1)$.

\section{Basic definitions}\label{sec:basic-defs}
Let $S$ be a diagnostic domain with the transition diagram $T$,
and actual trajectory 
$W = \langle\sigma^{w}_0, a^{w}_0, \sigma^{w}_1, \ldots, 
a^{w}_{n-1}, \sigma^{w}_n\rangle$. A pair, $\langle\Gamma_n,O^{m}_n\rangle$,
where $\Gamma_n$ is the recorded history of $S$ up to moment $n$
and $O^{m}_n$ 
is a collection of observations made by the agent
between times $n$ and $m$, will be called a {\em configuration}.
We say that a configuration
\beq\label{sympt}
{\cal S} = \langle \Gamma_{n},O^{m}_n \rangle
\end{equation}
is a {\em symptom} of the system's malfunctioning if
$\Gamma_{n}$ is consistent (w.r.t.~S) and $\Gamma_{n} \cup O^{m}_n$ is not.
Our definition of a candidate diagnosis of symptom (\ref{sympt}) 
is based on the notion of 
{\em explanation} from \cite{bg00}. According to that terminology,
an explanation, $E$, of symptom (\ref{sympt}) 
is a collection of statements
\beq\label{exp}
E = \{hpd(a_i,t) : 0 \leq t < n \mbox{ and } a_i \in A_e\}
\end{equation}
such that $\Gamma_{n} \cup O^{m}_n \cup E$ is consistent.

\begin{definition}\label{d1}
{\rm A {\em candidate diagnosis} $D$ of  symptom 
(\ref{sympt}) consists of an explanation $E(D)$ of (\ref{sympt}) 
together with the set $\Delta(D)$ of components of $S$ which
could possibly be damaged by 
actions from $E(D)$.  More precisely, 
$\Delta(D) = \{ c : M \models h(ab(c),m)\}$ 
for some model $M$ of $\Gamma_{n} \cup O^{m}_n \cup E(D)$.
}
\end{definition}

\begin{definition}\label{d2}
{\rm 
We say that $D$ is a  {\em diagnosis} of a symptom 
${\cal S}=\langle\Gamma_{n},O^{m}_n)$ if $D$ is a candidate diagnosis
of ${\cal S}$ in 
which all components in $\Delta$ are faulty, i.e.,
for any $c \in \Delta(D)$, $ab(c) \in \sigma^{w}_{m}$. 
}
\end{definition}

\section{Computing candidate diagnoses}\label{sec:computing}
In this section we show how the need for diagnosis can be determined
and candidate diagnoses found by the techniques of answer set 
programming \cite{mt99}. The proofs of the theorems presented here can be found
in Section \ref{app3}.

\st
From now on, we assume that we are given a diagnostic domain
$S=\langle \Sigma, T, W \rangle$. $SD$ will denote an action
description defining $T$.

\st
Consider a system description $SD$ of $S$ whose behavior up to the moment
$n$ from some interval $[0,N)$ is described by recorded history $\Gamma_n$.
(We assume that $N$ is sufficiently large for our application.)
We start by describing an encoding  of $SD$
into programs of A-Prolog suitable for execution by SMODELS \cite{ns97}.
Since SMODELS takes as an input programs with finite Herbrand
bases, references to lists should be eliminated from laws
of $SD$. To do that we expand 
the signature of $SD$ by new
terms - names of the corresponding statements of $SD$ - and consider a
mapping $\alpha$, from action descriptions of $\mathcal{AL}$ into programs of A-Prolog,
defined as follows:
\begin{enumerate}
\item $\alpha(causes(a,l_0,[l_1 \ldots l_m]))$ is the 
collection of atoms
\[
\begin{array}{l}
d\_law(d), head(d,l_0), action(d,a), \\prec(d,1,l_1), \ldots, prec(d,m,l_m), prec(d,m+1,nil)\mbox{.}
\end{array}
\]
Here and below $d$ will refer to the name of the corresponding law.
Statement $prec(d,i,l_i)$, with $1 \leq i \leq m$, says that $l_i$ 
is the $i$'th precondition of the law $d$; $prec(d,m+1,nil)$
indicates that the law has exactly $m$ preconditions.
This encoding of preconditions has a purely technical advantage.
It will allow us to concisely express the statements of the form `\emph{All
preconditions of a law $d$ are satisfied at moment $T$}'. (See rules (3-5)
in the program $\Pi$ below.)
\item $\alpha(caused(l_0,[l_1 \ldots l_m]))$ is the
collection of atoms
\[
\begin{array}{l}
s\_law(d), head(d,l_0), \\prec(d,1,l_1), \ldots, prec(d,m,l_m), prec(d,m+1,nil)\mbox{.}
\end{array}
\]
\item $\alpha(impossible\_if(a,[l_1 \ldots l_m]))$ is 
a constraint
$$
\begin{array}{lll}
\leftarrow &  h(l_1,T), \ldots,  h(l_n,T), & \\ 
           &  o(a,T)\mbox{.} &
\end{array}
$$

\noindent
where $o(a,t)$ stands for `{\em elementary action $a$ occurred at time $t$}'.
\end{enumerate}
By $\alpha(SD)$ we denote the result of applying $\alpha$ to
the laws of $SD$.
Finally, for any history, $\Gamma$, of $S$ 
$$\alpha(SD,\Gamma) = \Pi \cup \alpha(SD) \cup \Gamma$$
where $\Pi$ is defined as follows:
\[\Pi \: \left\{
\begin{array}{lll}
1\mbox{.} \ \ h(L,T^{\prime}) & \leftarrow & d\_law(D),\\
                     &            & head(D,L),\\
                     &            & action(D,A),\\
                     &            & o(A,T),\\
                     &            & prec\_h(D,T)\mbox{.}\\
2\mbox{.} \ \ h(L,T)          & \leftarrow & s\_law(D),\\
                     &            & head(D,L),\\
                     &            & prec\_h(D,T)\mbox{.}\\ 
3\mbox{.} \ \ all\_h(D,N,T)   & \leftarrow & prec(D,N,nil)\mbox{.} \\
4\mbox{.} \ \ all\_h(D,N,T)   & \leftarrow & prec(D,N,P), \\
                     &            & h(P,T), \\
                     &            & all\_h(D,N^\prime,T)\mbox{.} \\
5\mbox{.} \ \ prec\_h(D,T)    & \leftarrow & all\_h(D,1,T)\mbox{.}\\
6\mbox{.} \ \ h(L,T^\prime)   & \leftarrow & h(L,T),\\
                     &            & \no h(\overline{L}, T^{\prime})\mbox{.}\\
7\mbox{.} \ \          &        \leftarrow & h(L,T), h(\overline{L},T). \\
8\mbox{.} \ \ o(A,T)          & \leftarrow & hpd(A,T)\mbox{.}\\
9\mbox{.} \ \ h(L,0)          & \leftarrow & obs(L,0)\mbox{.}\\  
10\mbox{.}                & \leftarrow & obs(L,T),\\
                     &            & \no h(L,T)\mbox{.}       
\end{array}
\right.
\]
Here $D,A,L$ are variables for the names of 
laws, actions, and fluent literals  respectively, 
$T,T^{\prime}$ denote consecutive time points, 
and $N, N^{\prime}$ are variables for consecutive integers.
(To run this program under SMODELS we need to either define the above types
or add the corresponding typing predicates in the bodies of some rules of 
$\Pi$. These details will be omitted to save space.)
The relation $o$ is used instead of $hpd$ to distinguish
between actions observed ($hpd$), and actions hypothesized ($o$).

\st
Relation $prec\_h(d,t)$, defined by the rule (5) of $\Pi$,
says that all the preconditions of law $d$ are satisfied at moment $t$.
This relation is defined via an auxiliary relation $all\_h(d,i,t)$ (rules
(3), (4)), which
holds if the preconditions $l_i,\dots,l_m$ of $d$ are satisfied at moment
$t$. (Here $l_1,\dots,l_m$ stand for the ordering of preconditions of $d$
used by the mapping $\alpha$.)
Rules (1),(2) of $\Pi$ describe the effects of causal laws and constraints
of $SD$. Rule (6) is the inertia axiom \cite{mcc69}, rule (7) rules out inconsistent
states, rules (8) and (9) establish the relationship between
observations and the basic relations of $\Pi$, and rule (10), called the \emph{reality check},
guarantees that observations do not contradict the agent's expectations.

\st
(One may be tempted to replace ternary relation $prec(D,N,P)$ by
a simpler binary relation $prec(D,P)$ and to define
relation $prec\_h$ by the rules:
\[
\begin{array}{lcl}
\neg prec\_h(D,T) & \leftarrow & prec(D,P), \neg h(P,T)\mbox{.}\\
prec\_h(D,T) & \leftarrow & \no \neg prec\_h(D,T)\mbox{.}
\end{array}
\]
It is important to notice that this definition is incorrect since
the latter rule is defeasible and may therefore conflict with
the inertia axiom.) 

\st
The following terminology will be useful for describing the relationship
between answer sets of $\alpha(SD,\Gamma_n)$ and models of $\Gamma_n$.

\begin{definition}\label{def:defines}
Let $SD$ be an action description, and $A$ be a set of literals over
$lit(\alpha(SD,\Gamma_n))$. We say that $A$ \emph{defines} the sequence
\[
\tbeg \sigma_0, a_0, \sigma_1, \ldots, a_{n-1}, \sigma_{n} \tend
\]
if $\sigma_k = \{ l
\,\,|\,\, h(l,k) \in A\}$ and $a_k = \{ a \,\,|\,\, o(a,k) \in A\}$.
\end{definition}

\st
The following  theorem  establishes the relationship between 
the theory of actions in ${\cal AL}$ and logic programming. 
\begin{theorem}\label{th1}
If the initial situation of $\Gamma_n$ is {\em complete}, i.e. for any fluent
$f$ of $SD$, $\Gamma_n$ contains $obs(f,0)$ or $obs(\neg f,0)$
then
$M$ is a model of $\Gamma_n$ iff $M$ is defined by some answer set of 
$\alpha(SD,\Gamma_n)$.
\end{theorem}
(The theorem is similar to the result from \cite{tur97} which 
deals with a different language and uses the definitions from
\cite{mt95}.)

\st
Now let ${\cal S}$ be a configuration of the form (\ref{sympt}),
and let
\beq\label{e3}
Conf({\cal S})= \alpha(SD,\Gamma_n) \cup O^m_{n}\cup R
\end{equation}
where
\[R \: \left\{
\begin{array}{lll}
h(f,0) & \leftarrow & \no h(\neg f,0)\mbox{.}\\
h(\neg f,0) & \leftarrow & \no h(f,0)\mbox{.}
\end{array}
\right.
\]
for any fluent $f \in F$.
The rules of $R$ are sometimes called the {\em awareness axioms}. 
They guarantee that initially the agent considers all possible values
of the domain fluents.
(If the agent's information about the
initial state of the system is complete these axioms can be omitted.)
The following corollary forms the basis for
our diagnostic algorithms.
\begin{corollary}\label{c1}
Let ${\cal S} = \langle \Gamma_n,O^m_n\rangle$ where $\Gamma_n$
is consistent. Then configuration 
${\cal S}$ is a symptom of system's malfunctioning
iff program $Conf({\cal S})$ has no answer set.
\end{corollary}
To diagnose the system, $S$, we construct a program, $DM$, 
defining an {\em explanation space}  of our diagnostic agent - a collection
of sequences of exogenous events which could happen (unobserved) in the
system's past and serve as possible explanations
of unexpected observations. We call such programs {\em diagnostic modules}
for $S$.
The simplest diagnostic module, $DM_0$, is defined by rules:
\[DM_0 \: \left\{
\begin{array}{lll}
 o(A,T) & \leftarrow & 0 \leq T < n, \ x\_act(A),\\
             &            & \no \neg o(A,T)\mbox{.}\\
             &            &     \\
 \ \neg o(A,T) & \leftarrow & 0 \leq T < n, \ x\_act(A),\\
                  &            & \no o(A,T)\mbox{.}     
\end{array}
\right.
\]
or, in the more compact, {\em choice rule}, notation of SMODELS \cite{s99}

\st
\[
\begin{array}{lll}
\{o(A,T) : x\_act(A)\} & \leftarrow & 0 \leq T < n\mbox{.}
\end{array}
\]

\st
(Recall that a choice rule has  the form
$$m\{p(\overline{X}) \ :\ q(\overline{X})\}n \leftarrow body$$
and says that, if the body is satisfied by an answer set $AS$ of a program
then $AS$ must contain between $m$ and $n$ atoms of the form $p(\overline{t})$ 
such that $q(\overline{t}) \in AS$. For example, program
\[
\begin{array}{l}
\{ p(X) \,\,:\,\, q(X) \}\mbox{.} \\
q(a)\mbox{.}
\end{array}
\]
has two answer sets: $\{ q(a) \}$, and $\{ p(a), q(a) \}$.)

\st
Finding candidate diagnoses of symptom ${\cal S}$ can 
be reduced to finding answer sets of a {\em diagnostic program}
\beq\label{e4}
D_0({\cal S}) = Conf({\cal S}) \cup DM_0\mbox{.}
\end{equation}

\st
The link between answer sets and candidate diagnoses is described by the
following definition. 
\begin{definition}\label{def:D0-diagnosis}
Let $SD$ be a system description, 
$\mathcal{S}=\tbeg  \Gamma_n, O^m_n \tend$ be
a symptom of the system's malfunctioning,
$X$ be a set of ground literals,
and $E$ and $\delta$ be sets
of ground atoms.
We say that $\langle E, \Delta \rangle$ \emph{is determined by} $X$ if
\[
E=\{hpd(a,t) \,\,|\,\, o(a,t) \in X \land a \in A_e \}, \mbox{ and}
\]
\[
\Delta=\{ c \,\,|\,\, obs(ab(c), m) \in X \}\mbox{.}
\]
\end{definition}

\begin{theorem}\label{theorem:candidate_diag}
Let $\langle \Sigma, T, W \rangle$ be a diagnostic domain, $SD$ be a system
description of $T$, $\mathcal{S}=\tbeg  \Gamma_n, O^m_n \tend$ be
a symptom of the system's malfunctioning, and $E$ and $\delta$ be sets
of ground atoms. Then,
\[
\langle E, \Delta \rangle \mbox{ is a candidate diagnosis of } \mathcal{S}
\]
iff
\[
\langle E,\Delta \rangle \mbox{ is determined by an answer set of }
D_0(\mathcal{S})\mbox{.}
\]
\end{theorem}
The theorem justifies the following simple algorithm for computing
candidate diagnosis of a symptom ${\cal S}$:

\noindent
\begin{tabbing}
iiii\=iiii\=iiii\=iiii\=iiii\=iiii\=iiii\=iiii\=iiii\kill
{\bf function} $Candidate\_Diag($ ${\cal S}$: {\bf symptom} $)$;\\
\>{\bf Input}: a symptom ${\cal S}=\langle \Gamma_n,O^m_n \rangle$. \\
\>{\bf Output}: a candidate diagnosis of the symptom, or $\langle \emptyset,
\emptyset \rangle$ if no candidate \\
\>\>\>      diagnosis could be found. \\
\>{\bfseries var}\>\>$E$ : {\bfseries history}; \\
\>\>\>$\Delta$ : {\bfseries set of components}; \\
\>{\bfseries if} $D_0(\mathcal{S})$ is consistent {\bfseries then} \\
\>\>select an answer set, $X$, of $D_0(\mathcal{S})$; \\
\>\>compute $\langle E, \Delta \rangle$ determined by $X$; \\
\>\>{\bf return} $(\langle E,\Delta \rangle)$; \\
\>{\bfseries else} \\
\>\>$E := \emptyset$; $\Delta := \emptyset$; \\
\>{\bf return} $(\emptyset,\emptyset)$. \\
\>{\bfseries end} \\
\end{tabbing}

\st
Given a symptom $\mathcal{S}$, the algorithm constructs the program
$D_0(\mathcal{S})$ and passes it as an input to SMODELS \cite{ns97}, DLV
\cite{dlv}, DeReS \cite{deres}, or some
other answer set finder. If no answer set is found the algorithm returns 
$\langle \emptyset, \emptyset \rangle$. Otherwise the algorithm
returns a pair $\langle E, \Delta \rangle$ extracted from some answer 
set $X$ of the program. By Theorem \ref{theorem:candidate_diag} the
pair is a candidate diagnosis of $\mathcal{S}$. Notice that the set $E$
extracted from an 
answer set $X$ of $D_0(\mathcal{S})$ cannot be empty and hence the answer
returned by the function is unambiguos. (Indeed, using the Splitting Set
Theorem \cite{lif94a,turner96a} we can show that the existence of answer set of $D_0(\mathcal{S})$
with empty $E$ will lead to existence of an answer set of
$Conf(\mathcal{S})$, which, by Corollary \ref{c1}, contradicts to
$\mathcal{S}$ being a symptom.)
The algorithm can be illustrated by the following example. 
\begin{example}\label{ex1a}
{\rm
Let us again consider system 
$S$ from Example \ref{ex1}.
According to $\Gamma_1$ initially the switches $s_1$ and $s_2$ are open, 
all circuit components are ok, $s_1$ is 
closed by the agent, and {\em b} is protected. It is predicted that
{\em b} will be {\em on} at 1.  
Suppose that, instead, the agent observes that at time 1 
bulb {\em b} is {\em off}, i.e. 
$O_1 = \{obs(\neg on(b),1) \}$.
Intuitively, this is viewed as a symptom 
${\cal S}_0 = \langle \Gamma_1,O_1\rangle$
of malfunctioning of $S$.  By running SMODELS on 
$Conf({\cal S}_0)$
we discover that this program has no answer sets and therefore, by
Corollary \ref{c1},  ${\cal S}_0$ is indeed a symptom.
Diagnoses of ${\cal S}_0$ can be found by running SMODELS on 
$D_0({\cal S}_0)$ and extracting the necessary information
from the computed answer sets.
It is easy to check that, as expected, there are three candidate 
diagnoses:
\[
\begin{array}{lll}
D_1 = \langle\{o(brk,0)\},\{b\}\rangle&&\\
D_2 = \langle\{o(srg,0)\}, \{r\}\rangle&&\\
D_3 = \langle\{o(brk,0), o(srg,0)\}, \{b,r\}\rangle&&\\
\end{array}
\]
which corresponds to our intuition. Theorem \ref{th1} guarantees
correctness of this computation.
}
\end{example}
The basic diagnostic module $D_0$ can be modified
in many different ways. For instance, a simple modification, 
$D_1({\cal S})$,
which eliminates some candidate diagnoses containing actions unrelated
to the corresponding symptom can be constructed as follows. First, let us
introduce some terminology. Let $\alpha_i(SD)$ be a function that maps each
impossibility condition of $SD$ into a collection of atoms
\[
imp(d),action(d,a),prec(d,m+1,nil),prec(d,1,l_1),\ldots,prec(d,m,l_m),
\]
where $d$ is a new constant naming the condition, and $a$, $l_i$'s are
arguments of the condition. Let also $REL$ be the
following program:
\[REL \: \left\{
\begin{array}{llll} 
1\mbox{.}\  &rel(A,L) & \leftarrow & d\_law(D),\\
  &        &            & head(D,L),\\
  &       &            & action(D,A)\mbox{.}\\
2\mbox{.}\  & rel(A,L) & \leftarrow & law(D),\\
  &       &            & head(D,L),\\
  &       &            & prec(D,N,P),\\
  &       &            & rel(A,P)\mbox{.}\\
3\mbox{.}\  & rel(A_2,L) & \leftarrow & rel(A_1,L),\\
  &       &            & imp(D),\\
  &       &            & action(D,A_1),\\
  &       &            & prec(D,N,P),\\
  &       &            & rel(A_2,\overline{P})\mbox{.}\\
4\mbox{.} \  & rel(A) & \leftarrow & obs(L,T),\\
  &        &            & T \geq n,\\
  &        &            & rel(A,L)\mbox{.}\\
5\mbox{.}\   &     & \leftarrow & T < n,\\
   &     &            & o(A,T),\\
  &       &            & x\_act(A),\\
   &     &            & \no hpd(A,T),\\
  &     &            & \no rel(A)\mbox{.}
\end{array}
\right.
\]
and 
$$DM_1 = DM_0 \cup REL \cup \alpha_i(SD)\mbox{.}$$

\st
The new diagnostic module, $D_1$ is defined as
$$D_1({\cal S}) = Conf({\cal S}) \cup DM_1\mbox{.}$$
(It is easy to see that this modification is {\em safe},
i.e. $D_1$ will not miss any useful
predictions about the malfunctioning components.)
The difference between $D_0({\cal S})$ and $D_1({\cal S})$
can be seen from the following example.
\begin{example}\label{ex1b}
{\rm
Let us expand the system $S$ from Example \ref{ex1}
by a new component, $c$, unrelated to the circuit,
and an exogenous action $a$ 
which damages this component. It is easy to see
that diagnosis ${\cal S}_0$  from Example \ref{ex1}
will still be a symptom of malfunctioning
of a new system, $S_a$, and that the basic diagnostic module  
applied to $S_a$ will return diagnoses $(D_1) - (D_3)$ from
Example \ref{ex1a} together with new diagnoses containing $a$
and $ab(c)$, e.g. 
$$D_4 = \langle \{o(brks,0), o(a,0)\}, \{b,c\}\rangle.$$
Diagnostic module $D_1$ will ignore actions
unrelated to ${\cal S}$ and return only $(D_1) - (D_3)$. 
}
\end{example}
It may be worth noticing that the distinction between $hpd$ and $o$ allows
exogenous actions, including those unrelated to observations, 
to actually happen in the past.
Constraint (5) of program $REL$ only prohibits generating such actions in our
search for diagnosis.

\st
There are many other ways of improving quality of candidate diagnoses
by eliminating some redundant or unlikely diagnoses, and by ordering the
corresponding search space.
For instance, even more unrelated actions can be eliminated from the search space of 
our diagnostic modules by considering relevance relation $rel$
depending on time. This can be done by a simple modification of program REL
which is left as an exercise to the reader.
The diagnostic module $D_1$ can also be further modified
by limiting its search to recent occurrences of exogenous
actions. This can be done by  
$$D_2({\cal S}) = Conf({\cal S}) \cup DM_2$$ 
where $DM_2$ is obtained by replacing atom
$0 \leq T < n$ in the bodies of rules of $DM_0$ by
$n - m \leq T <n$.
The constant $m$ determines the time interval 
in the past that an agent is 
willing to consider in its search for possible explanations.
To simplify our discussion in the rest of the paper we {\em assume that} 
$m = 1$. Finally, the rule 
$$
\begin{array}{ll}
\leftarrow & k \{o(A,n-1)\}\mbox{.}\\
\end{array}
$$
added to $DM_2$ will eliminate all diagnoses containing
more than $k$ actions.
Of course the resulting module $D_3$ as well as $D_2$
can miss some candidate diagnoses and deepening of the search and/or increase
of $k$ may be necessary
if no diagnosis of a symptom is found. There are many other interesting
ways of constructing efficient diagnostics modules. We are especially
intrigued by the possibilities of using new features of answer sets 
solvers such as weight rules and minimize of SMODELS and weak constraints
of DLV \cite{dlv,dlv97} to specify a preference relation on diagnoses.
This however is a subject of further investigation.

\section{Finding a diagnosis}\label{sec:finding}
Suppose now the diagnostician has a candidate diagnosis $D$
of a symptom ${\cal S}$. Is it indeed a diagnosis?
To answer this question the agent should be able to test
components of $\Delta(D)$. Assuming that {\em no exogenous actions 
occur during testing} a diagnosis can be found by 
the following simple algorithm, $Find\_Diag({\cal S})$: 

\noindent
\begin{tabbing}
iiii\=iiii\=iiii\=iiii\=iiii\=iiii\=iiii\=iiii\=iiii\kill
{\bf function} $Find\_Diag($ {\bf var} ${\cal S}$: {\bf symptom} $)$;\\
\>{\bf Input}: a symptom ${\cal S}=\langle \Gamma_n,O^m_n \rangle$. \\
\>{\bf Output}: a diagnosis of the symptom, or $\langle \emptyset, \emptyset \rangle$ if no
diagnosis \\
\>\>could be found. Upon successful termination of the loop, the set
$O^m_n$\\
\>\>is updated in order to incorporate the results of the tests\\ 
\>\> done during the search for a diagnosis. 
\\
\>{\bf var}\>\>$O$, $E$ : {\bf history}; \\
\>\>\>$\Delta$, $\Delta_0$ : {\bf set of components}; \\
\>\>\>$diag$ : {\bf bool}; \\
\>$O := O^m_n;$ \\
\>{\bf repeat} \\
\>\>$\langle E,\Delta \rangle :=  Candidate\_Diag( \,\, \langle \Gamma_n,O \rangle \,\, );$\\
\>\>{\bf if} $E = \emptyset$ \{ no diagnosis could be found \} \\
\>\>\>{\bf return}($\langle E, \Delta \rangle$); \\
\>\>$\mbox{diag} := true; \ \ \ $ $ \Delta_0 := \Delta$;\\
\>\>{\bf while} $\ \Delta_0 \not= \emptyset\ $ {\bf and} $\mbox{ diag }$ {\bf do}\\
\>\>\>select $c \in \Delta_0; \ \ \ $ $\Delta_0 := \Delta_0 \setminus \{c\}$;\\
\>\>\> {\bf if} $\ observe(m,ab(c))=ab(c)\ $ {\bf then}\\
\>\>\>\>$O := O \cup obs(ab(c),m);$\\
\>\>\>{\bf else}\\
\>\>\>\>$O := O \cup obs(\neg ab(c),m);$\\
\>\>\>\>$\mbox{diag} := false$;\\
\>\>\>{\bf end} \\
\>\>{\bf end} \{while\}\\
\>{\bf until} $\mbox{diag}$;\\
\>$O^m_n := O$; \\
\>{\bf return} $(\langle E,\Delta \rangle)$.
\end{tabbing}
%

\noindent
The properties of $Find\_Diag$ are described by the following theorem.

\begin{theorem}\label{theorem:find_diag}
Let $\langle \Sigma, T, W \rangle$ be a diagnostic domain, $SD$ be a
system description of $T$, and $\mathcal{S}=\tbeg  \Gamma_n, O^m_n \tend$
be a symptom of the system's malfunctioning. Then,
\begin{enumerate}
\item
$Find\_Diag(\mathcal{S})$ terminates;
\item
let $\tbeg E, \Delta \tend = Find\_Diag(\mathcal{S})$, where the value of
variable $\mathcal{S}$ is set to $\mathcal{S}_0$. If $\Delta \not= \emptyset$,
then
\[
\tbeg E, \Delta \tend \mbox{ is a diagnosis of } \mathcal{S}_0;
\]
otherwise, $\mathcal{S}_0$ has no diagnosis.
\end{enumerate}
\end{theorem}

%
%

\noindent
To illustrate the algorithm, consider the following example.
\begin{example}\label{ex3}
{\rm 
Consider the system $S$ from Example \ref{ex1} and a 
history $\Gamma_1$ in which
{\em b} is not protected,  
all components of $S$ are ok, both switches are open, and the agent closes 
$s_1$ at time 0.  At time 1, he  observes that 
the bulb {\em b} is not lit, considers
${\cal S} = \langle \Gamma_1, O_1\rangle$ where
$O_1 = \{obs(\neg on(b),1)\}$ and calls
function $Need\_Diag({\cal S})$
which searches for an answer set of $Conf({\cal S})$. There are no such sets,
the diagnostician realizes he has a symptom to diagnose and calls 
function $Find\_Diag({\cal S})$.
Let us assume that the first call to  $Candidate\_Diag$ 
returns 
$$PD_1 = \langle \{o(srg,0)\},\{r,b\} \rangle$$
Suppose that the agent selects component $r$ from $\Delta$  
and determines that it is not faulty.
Observation $obs(\neg ab(r),1)$ will be added to $O_1$, $diag$ will be set to $false$
and the program will call
$Candidate\_Diag$ again with the updated symptom ${\cal S}$ as a parameter. 
$Candidate\_Diag$ will return another possible diagnosis 
$$PD_2 = \langle \{o(brk,0)\},\{b\} \rangle$$
The agent will test bulb {\em b}, find it 
to be faulty, add observation 
$obs(ab(b),1)$ to $O_1$ and  return $PD_2$.
If, however, according to our actual trajectory, $W$, the bulb is still ok, the
function returns $\langle \emptyset, \emptyset \rangle$. No diagnosis is found
and the agent (or its designers) should start looking for a modeling error.
}
\end{example}
\section{Diagnostics and repair}\label{sec:repair}
Now let us consider a scenario which is only slightly different from that
of the previous example.
\begin{example}\label{ex4}
{\rm 
Let  $\Gamma_1$ and observation
$O_1$ be as in Example \ref{ex3} and
suppose that the program's first call to
$Candidate\_Diag$ returns 
$PD_2$, {\em b} is found to be faulty, 
$obs(ab(b),1)$ is added to $O_1$, and $Find\_Diag$ 
returns $PD_2$.
The agent proceeds to have {\em b} repaired but, to his 
disappointment, discovers that {\em b} is still not on!
Intuitively this means that $PD_2$ is a wrong diagnosis
 - there must have been a power surge at 0.
}
\end{example}
For simplicity we assume that,
similar to testing,
repair occurs in well controlled
environment, i.e. {\em no exogenous actions happen during the repair process}.
The example shows that,
{\em in order to find  a correct explanation
of a symptom, it is essential for an agent 
to repair damaged components and observe the behavior of the system
after repair}.  
To formally model this process we  introduce a special action, 
$repair(c)$, for every component $c$ of $S$. 
The effect of this action will be defined
by the causal law:
$$causes(repair(c),\neg ab(c),[])$$ 
The diagnostic process will be now modeled by the following algorithm:
(Here ${\cal S} = \langle\Gamma_n,O^m_n\rangle$ and
$\{obs(f_i,k)\}$ is a collection of observations
the diagnostician makes to test his repair at moment $k$.) 
\begin{tabbing}
iiii\=iiii\=iiii\=iiii\=iiii\=iiii\=iiii\=iiii\=iiii\kill
{\bf function} $Diagnose({\cal S})$ : {\bf boolean};\\
\>{\bf Input}: a symptom ${\cal S}=\langle \Gamma_n,O^m_n \rangle$. \\
\>{\bf Output}: {\em false} if no diagnosis can be found.
Otherwise\\
\>\> repairs the system, updates $O_{n}^{m}$, and returns {\em true}.\\
\>{\bf var}\>\>$E$ : {\bf history}; \\
\>\>\>$\Delta$ : {\bf set of components}; \\
\>$E=\emptyset$; \\
\>{\bf while} $Need\_Diag(\langle \Gamma_n \cup E,O^m_n \rangle )$ {\bf do}\\
\>\>\>$\langle E,\Delta\rangle = Find\_Diag(\langle \Gamma_n,O^m_n \rangle );$\\
\>\>\>{\bf if} $E = \emptyset$ {\bf then} {\bf return}(false)\\ 
\>\>\>{\bf else}\\
\>\>\>\>$Repair(\Delta)$;\\
\>\>\>\>$O^m_n := O^m_n \cup \{hpd(repair(c),m) : c \in \Delta\}$; \\
\>\>\>\>$m := m+1$;\\
\>\>\>\>$O^m_n := O^{m-1}_n \cup \{obs(f_i,m)\}$;\\
\>\>\>{\bf end}\\
\>{\bf end}\\
\> {\bf return}(true);
\end{tabbing}
\begin{example}\label{ex5}
{\rm 
To illustrate the above algorithm let us go back to the agent from 
Example \ref{ex4}
who just discovered diagnosis $PD_2 = \langle \{o(brk,0)\},\{b\}
\rangle$. He will repair
the bulb and check if the bulb is lit. It is not, and therefore a new
observation is recorded as follows:
$$O_1 := O_1 \cup \{hpd(repair(b),1),obs(\neg on(b),2)\}$$
$Need\_Diag({\cal S})$ will detect a continued  need for diagnosis,
$Find\_Diag({\cal S})$ will
return $PD_1$, which, after new repair and testing will
hopefully prove to be the right diagnosis. 
}
\end{example}
The diagnosis produced by the above algorithm can be viewed as a 
reasonable interpretation of discrepancies between the agent's predictions
and actual observations. To complete our analysis of step 1 of
the agent's acting and reasoning loop we need to explain how
this interpretation can be incorporated in the agent's history.
If the diagnosis discovered is unique then the answer
is obvious - $O$ is simply added to $\Gamma_n$. If however 
faults of the system components can be caused by different sets of
exogenous actions the situation becomes more subtle.
Complete investigation of the issues involved is the subject of further
research.

\section{Related work}\label{sec:rel-work}
There is a numerous collection of papers on diagnosis
many of which substantially influenced the author's views
on the subject.
The roots of our approach go back to \cite{r87}
where diagnosis for a static environment were formally defined in logical
terms. To the best of our knowledge the first published
extensions of this work to dynamic domains appeared in \cite{thi97},
where dynamic domains were described in fluent calculus \cite{thi98}, and in
\cite{mi97} which used situation calculus \cite{mcc69}. 
Explanation of malfunctioning of system components
in terms of unobserved exogenous actions was first
clearly articulated in \cite{mi98}. Generalization and extensions
of these ideas \cite{bms00} which specifies dynamic domains in 
action language ${\cal L}$, can be viewed as a starting point of the
work presented in this paper.
The use of a simpler action language ${\cal AL}$ allowed us to
substantially simplify the basic definitions of \cite{bms00}
and to reduce the computation of diagnosis to finding stable models
of logic programs. As a result we were able to incorporate diagnostic
reasoning in a general agent architecture based on the answer set programming
paradigm, and to  combine diagnostics with planning and other activities of
a reasoning agent. On another hand \cite{bms00} addresses some questions which
are not fully addressed by our paper. In particular, the underlying action
language of \cite{bms00} allows non-deterministic and knowledge-producing
actions absent in our work. While our formulation allows immediate
incorporation of the former, incorporation of 
the latter seems to substantially increase conceptual complexity of the
formalism. This is of course the case in \cite{bms00} too but
we believe that the need for such increase in complexity remains an open
question. Another interesting related work is \cite{o00}. In this paper the 
authors address the problem of dynamic diagnosis using the notion of
pertinence logic from \cite{oc99}. The formalism allows to define dynamic diagnosis
which, among other things, can model  intermittent faults of the system. 
As a result it provides a logical account of the following scenario:
Consider a person trying to shoot a turkey. 
Suppose that the gun is initially loaded, the agent shoots, observes
that the turkey is not dead, and shoots one more time. Now the turkey is dead.
The pertinence formalism of \cite{o00} does not claim
inconsistency - it properly determines that the gun has an intermittent
fault. Our formalism on another hand is not capable of modeling this scenario -
to do that we need to introduce non-deterministic actions.
Since, in our opinion, the use of pertinence logic substantially 
complicates action formalisms it is interesting to see if such use
for reasoning with intermittent faults can always be avoided
by introducing non-determinism.
Additional comparison of the action languages based
approach to diagnosis with other related approaches can be found in
\cite{bms00}.

Finally, let us mention that the reasoning algorithms proposed in this paper
are based on recent discoveries of close
relationship between 
A-Prolog and reasoning about effects of actions \cite{mt95}
and the ideas from answer set programming 
\cite{mt99,n99,l99}.
This approach of course would be impossible without existence
of efficient answer set reasoning systems.
The integration of diagnostics and other activities
is based on the agent architecture from \cite{bg00}.

\section{Conclusions and further work}\label{sec:conclusions}
The paper describes a work on the development of a diagnostic 
problem solving agent in which a mathematical model of an agent
and its environment is based on the theory of action 
language ${\cal AL}$ from \cite{bg00}. The language, which contains the means
for representing concurrent actions and fairly complex relations 
between fluents, is used to give  concise
descriptions of transition diagrams characterizing possible
trajectories of the agent domains as well as the domains' recorded
histories. 
In this paper we:

\begin{itemize}
\item
Establish a  close relationship between ${\cal AL}$
and logic programming under the answer set semantics
which allows reformulation of the agent's knowledge in A-Prolog.  
These results build on previous work connecting action languages
and logic programming.

\item
Give definitions of symptom, candidate diagnosis,
and diagnosis which we believe to be simpler 
than similar definitions we were able to find in the literature.

\item
Suggest a new algorithm for computing candidate 
diagnoses. (The algorithm is based on answer set programming and views 
the search for candidate diagnoses as `planning in the past'.)

\item
Suggest some simple ways of using A-Prolog to declaratively
limit the diagnostician's search space. This leads to higher quality
diagnosis and substantial improvements in the diagnostician's efficiency.

\item
Give a simple account of diagnostics, testing and repair
based on the use of answer set solvers. The resulting algorithms, which
are shown to be provenly correct, can be easily incorporated in the agent's
architecture from \cite{bg00}.

\end{itemize}

\st
In our further work we plan to:

\begin{itemize}
\item
Expand our results to more
expressive languages, i.e. those with non-deterministic actions,
defeasible causal laws, etc. 

\item
Find more powerful declarative ways of limiting the diagnostician's
search space. This can be done by expanding A-Prolog by ways of expressing
preferences between different rules or by having the agent plan
observations aimed at eliminating large clusters of possible diagnosis.
In investigating these options we plan to build on related work
in \cite{dlv97} and \cite{bms00,ss00}.

\item
Test the efficiency of the suggested algorithm on medium size
applications. 

\end{itemize}


\section{The syntax and semantics of A-Prolog}\label{app1}
In this section we give a brief introduction to the syntax
and semantics of a comparatively simple variant of A-Prolog. 
The syntax of the language is 
determined by a signature $\Sigma$ consisting of types,
$types(\Sigma) = \{\tau_0,\dots,\tau_m\}$, object constants\\ 
$obj(\tau,\Sigma) = \{c_0,\dots,c_m\}$ for each type $\tau$, and 
typed function and predicate constants $func(\Sigma)=\{f_0,\dots,f_k\}$ and 
$pred(\Sigma) = \{p_0,\dots,p_n\}$.
We will assume that the signature contains symbols for integers and
for the standard relations of arithmetic. 
Terms are built as in typed first-order languages;  
positive literals (or atoms) have the form $p(t_{1},\ldots,t_{n})$,  
where $t$'s are terms of proper types and $p$ is a predicate symbol of
arity $n$; negative literals are of the form $\neg 
p(t_{1},\ldots,t_{n})$. In our further discussion we often
write $p(t_1,\dots,t_n)$ as $p(\overline{t})$.
The symbol $\neg$ is called {\em classical} or {\em 
strong} negation.
Literals of the form $p(\overline{t})$ 
and $\neg p(\overline{t})$ are called contrary. By 
$\overline{l}$ we denote a literal contrary to $l$. 
Literals and terms not containing variables are called {\em  ground}. 
The sets of all ground terms, atoms and literals over $\Sigma$ will  
be denoted by $terms(\Sigma)$, $atoms(\Sigma)$ and $lit(\Sigma)$ respectively. 
For a set $P$ of predicate symbols from $\Sigma$,  
$atoms(P,\Sigma)$ ($lit(P,\Sigma)$)  
will denote the sets of ground atoms (literals) of $\Sigma$ formed  
with predicate symbols from $P$. Consistent sets of ground  
literals over signature $\Sigma$, containing all arithmetic literals 
which are true  under the standard interpretation of their symbols,
are called  {\em states} of $\Sigma$ and denoted
by $states(\Sigma)$.
 
\st 
A rule of A-Prolog is an expression of the form  
\beq\label{eq1} 
l_0 \leftarrow l_1, \dots ,l_m,\no l_{m+1}, \dots, \no l_n  
\end{equation} 
where $n \geq 1$, $l_i$'s are literals, $l_0$ is a literal or the symbol
$\perp$, 
and $\no$ is a  logical 
connective called {\em negation as failure} or {\em default negation}. 
An expression
$\no l$ says that there is no reason to believe in $l$.
An {\em extended literal} is an expression of the form $l$ or 
$\no l$ where $l$ is a literal.
A rule (\ref{eq1}) is called a {\em constraint} if $l_0 = \perp$. 
 
\st 
Unless otherwise stated, we  assume that the $l's$ in  
rules (\ref{eq1}) are ground. 
Rules with variables (denoted by capital letters) will be used 
only as a shorthand for the sets of their ground instantiations. 
This approach is justified for the so
called closed domains, i.e.~domains satisfying 
the domain closure assumption \cite{rei78} which 
asserts that {\em all objects in the domain of discourse have names in
the language of $\Pi$}.

\noindent 
A pair $\langle \Sigma,\Pi \rangle$ where $\Sigma$ is a signature and $\Pi$ is 
a collection of rules over $\Sigma$ is called a {\em logic 
program}. (We often denote such pair by its second element 
$\Pi$. The corresponding signature will be denoted by $\Sigma(\Pi)$.)

 
\st
We say that a literal $l \in lit(\Sigma)$ is $true$ in a
state $X$ of $\Sigma$ if $l \in X$; 
$l$ is $false$ in $X$ if $\overline{l} \in X$;
Otherwise, $l$ is unknown. $\perp$ is false in $X$.

\st
Given a signature $\Sigma$ and a set of predicate symbols $E$, $lit(\Sigma,E)$
denotes the set of all literals of $\Sigma$ formed by predicate symbols from
$E$. If $\Pi$ is a ground program, $lit(\Pi)$ denotes the set of all atoms occurring
in $\Pi$, together with their negations, and $lit(\Pi,E)$ denotes the set of all
literals occurring in $lit(\Pi)$ formed by predicate symbols from $E$.

\st 
The answer set semantics of a logic program $\Pi$ assigns to 
$\Pi$ a collection of {\em answer sets} -- consistent sets of ground literals 
over signature $\Sigma(\Pi)$ corresponding to beliefs 
which can be built by a rational reasoner on the basis of rules of $\Pi$.
In the construction of these beliefs the reasoner is assumed to be guided 
by the following informal principles:
\begin{itemize}
\item He should satisfy the rules of $\Pi$, understood as 
constraints of the form:
{\em If one believes in the body of a rule one must belief in its head}. 

\item He cannot believe in $\perp$ (which is understood as falsity).

\item He should adhere to the {\em rationality principle} which says that
{\em one shall not believe anything he is not forced to believe}.

\end{itemize}
The precise definition of answer sets
will be first given for programs whose rules do not contain default 
negation. Let $\Pi$ be such a program and let $X$ be a state of
$\Sigma(\Pi)$.
We say that $X$ is {\em closed} under $\Pi$ if, for every
rule $head \leftarrow body$ of $\Pi$, head is true in $X$ whenever $body$
is true in $X$. (For a constraint this condition means that the body is not 
contained in $X$.)
\begin{definition}\label{as}{(Answer set -- part one)}\\
{\rm  
A state $X$ of $\Sigma(\Pi)$
 is an {\em answer set} for $\Pi$ if $X$ is minimal 
(in the sense of set-theoretic inclusion) among
the sets closed under $\Pi$.  
}
\end{definition}
It is clear that a program without default negation
can have at most one answer set. 
To extend this definition to arbitrary programs, take any program $\Pi$,
and let $X$ be a state of $\Sigma(\Pi)$. The {\em reduct}, $\Pi^X$, of $\Pi$ 
relative to $X$ is the set of rules 
$$l_0 \leftarrow l_1,\dots,l_m$$
for all rules (\ref{eq1}) in $\Pi$ such that $l_{m+1},\dots,l_n \not\in X$.
Thus $\Pi^X$ is a program without default negation.
\begin{definition}\label{as2}{(Answer set -- part two)}\\
{\rm  
A state $X$ of $\Sigma(\Pi)$ is
an answer set for $\Pi$ if $X$ is an answer set for $\Pi^X$. 
}
\end{definition}
(The above definition differs slightly from the original definition in
\cite{gl91}, which allowed the inconsistent answer set, $lit(\Sigma)$. Answer
sets defined in this paper correspond to consistent answer sets of the original
version.)
%
%
%

\section{Syntax and semantics of the causal laws of \ALns}\label{app2}
An action description of \AL is a
collection of propositions of the form
\begin{enumerate}
\item $causes(a_e,l_0,[l_1,\dots,l_n])$,
\item $caused(l_0,[l_1,\dots,l_n])$, and
\item $impossible\_if(a_e,[l_1,\dots,l_n])$
\end{enumerate}
where $a_e$ is an elementary action
and $l_0,\dots,l_n$ are fluent literals from
$\Sigma$. The first proposition says that, if
the elementary action $a_e$ were to be executed in a situation in which
$l_1,\dots,l_n$ hold, the fluent literal $l_0$ will be caused to
hold in the resulting situation. Such propositions are called {\em
dynamic causal laws}. 
The second proposition, called a
{\em static causal law}, says that, in an arbitrary situation,
the truth of fluent literals, $l_1,\dots,l_n$ is sufficient to
cause the truth of $l_0$. The last proposition says that action
$a_e$ cannot be performed in any
situation in which $l_1,\dots,l_n$ hold.
(The one presented here is actually a
simplification of \ALns. Originally $impossible\_if$ took as
argument a compound action rather than an elementary one.
The restriction on $a_e$ being elementary is not
essential and can be lifted. We require it to simplify the presentation).
To define the transition diagram, $T$, given by 
an action description ${\cal A}$ of ${\cal AL}$
we use the
following terminology and notation. A set $S$ of fluent literals is
closed under a set $Z$ of static causal laws if $S$ includes the
head, $l_0$, of every static causal law  such that
$\{l_1,\dots,l_n\} \subseteq S$. The set $Cn_Z(S)$ of {\em
consequences} of $S$ under $Z$ is the smallest set of fluent literals that
contains $S$ and is closed under $Z$. 
$E(a_e,\sigma)$ stands for the set of all fluent literals $l_0$ for which
there is a dynamic causal law $causes(a_e,l_0,[l_1,\dots,l_n])$
in ${\cal A}$ such that $[l_1,\dots,l_n] \subseteq \sigma$.
$E(a,\sigma) = \bigcup_{a_e \in a}E(a_e,\sigma)$.
The transition system $ T = \langle
{\cal S},{\cal R}\rangle$ {\em described} by an action description
${\cal A}$ is defined as follows:
\begin{enumerate}
\item ${\cal S}$ is the collection of all complete and consistent sets
of fluent literals of $\Sigma$ closed under the static laws of
${\cal A}$,
\item $\cal R$ is the set of all triples
$\langle \sigma,a,\sigma^\prime \rangle$ such that ${\cal A}$ does not contain a
proposition of the form $impossible\_if(a,[l_1,\dots,l_n])$ such
that $[l_1,\dots,l_n] \subseteq \sigma$ and
\begin{equation}\label{eq:mccain-turner}
\sigma^\prime = Cn_Z(E(a,\sigma) \cup (\sigma \cap \sigma^\prime))
\end{equation}
where $Z$ is the set of all static causal laws of ${\cal A}$.
The argument of $Cn(Z)$ in
(\ref{eq:mccain-turner}) is the union of the set $E(a,\sigma)$ of the ``direct
effects'' of $a$ with the set $\sigma \cap \sigma^\prime$ of facts that are
``preserved by inertia''. The application of $Cn(Z)$ adds the
``indirect effects'' to this union.
\end{enumerate}
We call an action description {\em deterministic}
if for any state $\sigma_0$ and action $a$ there is
at most one such successor state $\sigma_1$.

\st
The above definition of $T$ is from \cite{mcc97} and  is the product of 
a long investigation of the nature of causality. (See for instance,
\cite{lif97b,th97}.) Finding this definition required the good understanding
of the nature of causal
effects of actions in the presence of complex interrelations between fluents.
An additional level of complexity is added by the need to specify what
is not changed by actions. The latter, known as the {\em frame problem},
is often
reduced to the problem of finding a concise and accurate representation of
the inertia axiom -- a default which says that {\em things normally stay as
they are} \cite{mcc69}. The search for such a representation substantially
influenced  AI research during the last twenty years.
An interesting account of history of this research 
together with some possible solutions can be found in \cite{sha97}.

\section{Properties of logic programs}\label{appendix:extension}
In this section we introduce several properties of logic programs which will be
used, in the next appendix, to prove the main theorem of this paper.

\st
We begin by summarizing two useful definitions from \cite{bd94}.

\begin{definition}
Let $q$ be a literal and $P$ be a logic program. The \emph{definition} of $q$ in
$P$ is the set of all rules in $P$ which have $q$ as their head.
\end{definition}

\begin{definition}[Partial Evaluation]
Let $q$ be a literal and $P$ be a logic program. Let
\[
\begin{array}{rcl}
q       & \hif  & \Gamma_1\mbox{.} \\
q       & \hif  & \Gamma_2\mbox{.} \\
        & \cdots &
\end{array}
\]
be the definition of $q$ in $P$. The Partial Evaluation of $P$ w.r.t. $q$
(denoted by $e(P,q)$) is the program obtained from $P$ by replacing every rule of
the form
\[
p \hif \Delta_1, q, \Delta_2\mbox{.}
\]
with rules
\[
\begin{array}{rcl}
p       & \hif  & \Delta_1, \Gamma_1, \Delta_2\mbox{.} \\
p       & \hif  & \Delta_1, \Gamma_2, \Delta_2\mbox{.} \\
        & \cdots &
\end{array}
\]
Notice that, according to Brass-Dix Lemma \cite{bd94}, $P$ and $e(P,q)$ are equivalent (written
$P \simeq e(P,q)$), i.e. they have the same answer sets.
\end{definition}

\st
The following expands on the results from \cite{bd94}.

\begin{definition}[Extended Partial Evaluation]\label{def:e}
Let $P$ be a ground program, and $\vec{q}=\tbeg q_1, q_2, \ldots, q_n \tend$ be a
sequence of literals. The Extended Partial Evaluation of $P$ w.r.t. $\vec{q}$
(denoted by $e(P,\vec{q})$) is defined as follows:
\begin{equation}\label{eq:e:1}
e(P,\vec{q}) = \left\{
\begin{array}{ll}
P & \mbox{if } n=0 \\
e(e(P,q_n),\tbeg q_1, q_2, \ldots, q_{n-1} \tend) & \mbox{otherwise}
\end{array}\right.
\end{equation}
From now on, the number of elements of $\vec{q}$ will be denoted by
$|\vec{q}|$.
\end{definition}

\begin{definition}[Trimming]\label{def:t}
Let $\vec{q}$ and $P$ be as above. The Trimming of $P$ w.r.t.
$\vec{q}$ (denoted by $t(P,\vec{q})$) is the program obtained by dropping the
definition of the literals in $\vec{q}$ from $e(P,\vec{q})$.
\end{definition}

\begin{lemma}\label{lemma:trans}
Let $P$ and $R$ be logic programs, such that
\begin{equation}\label{hp:trans:1}
P \simeq R\mbox{.}
\end{equation}
Then, for any sequence of literals $\vec{q}$,
\begin{equation}\label{th:trans:1}
e(P,\vec{q}) \simeq e(R,\vec{q})\mbox{.}
\end{equation}
\if T\extPaper
\Begproof
By induction on $|\vec{q}|$.

\st
Base case: $|\vec{q}|=0$. By Definition \ref{def:e}, $e(P,\vec{q})=P$ and
$e(R,\vec{q})=R$. Then, (\ref{hp:trans:1}) can be rewritten as
$e(P,\vec{q}) \simeq e(R,\vec{q})$.

\st
Inductive step: let us assume that (\ref{th:trans:1}) holds for
$|\vec{q}|=n-1$ and show that it holds for $|\vec{q}|=n$. Since $P
\simeq R$, by inductive hypothesis
\begin{equation}\label{eq:trans:1}
e(P,\tbeg q_1, \ldots, q_{n-1} \tend) \simeq e(R,\tbeg q_1, \ldots, q_{n-1}
\tend)\mbox{.}
\end{equation}
By the Brass-Dix Lemma,
\begin{equation}\label{eq:trans:1b}
e(P,q_n) \simeq P \mbox{ and } e(R,q_n) \simeq R.
\end{equation}
Again by inductive hypothesis, (\ref{eq:trans:1b}) becomes
$e(e(P,q_n),\tbeg q_1, \ldots, q_{n-1} \tend) \simeq 
e(P,\\\tbeg q_1, \ldots, q_{n-1} \tend)$ and similarly for $R$. Then, 
(\ref{eq:trans:1}) can be rewritten as
\begin{equation}\label{eq:trans:2}
e(e(P,q_n),\tbeg q_1,\ldots, q_{n-1} \tend) \simeq e(e(R,q_n),\tbeg q_1,\ldots,
q_{n-1} \tend)\mbox{.}
\end{equation}
By Definition \ref{def:e}, $e(e(P,q_n),\tbeg q_1,\ldots, q_{n-1}
\tend)=e(P,\vec{q})$ and similarly for $R$. Then, (\ref{eq:trans:2}) becomes
\[
e(P,\vec{q}) \simeq e(R,\vec{q})\mbox{.}
\]
\Endproof
\fi
\end{lemma}

\begin{lemma}\label{lemma:equiv}
Let $\vec{q}$ be a sequence of literals and $P$ be a logic program. Then,
\begin{equation}\label{th:equiv:1}
P \simeq e(P,\vec{q})\mbox{.}
\end{equation}
\if T\extPaper
\Begproof
By induction on $|\vec{q}|$.

\st
Base case: $|\vec{q}|=0$. $P=e(P,\vec{q})$ by Definition
\ref{def:e}.

\st
Inductive step: let us assume that (\ref{th:equiv:1}) holds for
$|\vec{q}|=n-1$ and show that it holds for $|\vec{q}|=n$.

\st
By the Brass-Dix Lemma, $P \simeq e(P,q_n)$. Then, Lemma \ref{lemma:trans} can be
applied to $P$ and $e(P,q_n)$, obtaining
\begin{equation}\label{eq:equiv:1}
e(P,\tbeg q_1,\ldots, q_{n-1} \tend) \simeq  e(e(P,q_n),\tbeg q_1,\ldots,
q_{n-1} \tend)\mbox{.}
\end{equation}
Since, by inductive hypothesis, $P \simeq e(P,\tbeg q_1, \ldots, q_{n-1}
\tend)$, (\ref{eq:equiv:1}) can be rewritten as
\[
P \simeq e(e(P,q_n),\tbeg q_1,\ldots, q_{n-1} \tend),
\]
which, by Definition \ref{def:e}, implies (\ref{th:equiv:1}).
\Endproof
\fi
\end{lemma}

\st
The following expands similar results from \cite{gs98}, making them suitable for
our purposes.

\begin{definition}[Strong Conservative Extension]\label{def:strongext}
Let $P_1$ and $P_2$ be ground programs such that $lit(P_1) \subseteq lit(P_2)$.
Let $Q$ be $lit(P_2) \setminus lit(P_1)$.

\st
We say that $P_2$ is a Strong Conservative Extension of $P_1$ w.r.t. $Q$ (and
write $P_2 \succ_Q P_1$) if:
\begin{itemize}
\item
if $A$ is an answer set of $P_2$, $A \setminus Q$ is an answer set of
$P_1$;
\item
if $A$ is an answer set of $P_1$, there exists a subset $B$ of $Q$ such
that $A \cup B$ is an answer set of $P_2$.
\end{itemize}
\end{definition}

\begin{lemma}\label{lemma:strongext}
Let $P$ be a ground program, $Q \subseteq lit(P)$, and $\vec{q}=\tbeg q_1,
\ldots, q_n \tend$ be an ordering of $Q$. If $Q \cap
lit(t(P,\vec{q})) = \emptyset$, then
\begin{equation}\label{th:strongext:1}
P \mbox{ is a Strong Conservative Extension of } t(P,\vec{q}) \mbox{ w.r.t
} Q.
\end{equation}
\if T\extPaper
\Begproof
Notice that, under the hypotheses of this lemma, the complement,
$\overline{Q}$, of $Q$ is a splitting set for $e(P,\vec{q})$, with
$bottom_{\overline{Q}}(e(P,\vec{q}))=t(P,\vec{q})$. Then, by the 
Splitting Set Theorem, and by Definition \ref{def:strongext},
\begin{equation}\label{eq:strongext:1}
e(P,\vec{q}) \succ_Q t(P,\vec{q})\mbox{.}
\end{equation}
By Lemma \ref{lemma:equiv}, $P \simeq e(P,\vec{q})$. Then,
(\ref{eq:strongext:1}) can be rewritten as
\[
P \succ_Q t(P,\vec{q})\mbox{.}
\]
\Endproof
\fi
\end{lemma}

\section{Proofs of the theorems}\label{app3}
\subsection{Proof of Theorem \ref{th1}}
The proof of Theorem \ref{th1} will be given in several steps.

\st
First of all, we
define a simplified encoding of an action description, $SD$, in
A-Prolog. Then, we prove that the answer sets of the programs generated using
this encoding correspond exactly to the paths in the transition diagram
described by $SD$.

\st
Later, we extend the new encoding and prove that, for every recorded history
$\Gamma_n$, the models of $\Gamma_n$ are in a one-to-one correspondence with the
answer sets of the programs generated by this second encoding.

\st
Finally, we prove that programs obtained by applying this second encoding are
essentially `equivalent' to those generated with the encoding presented in Section
\ref{sec:computing}, which completes the proof of Theorem \ref{th1}. In
addition, we present a corollary that extends the theorem to
the case in which the initial situation of $\Gamma_n$ is not complete.

\subsubsection{Step 1}
The following notation will be useful in our further discussion. Given a time
point $t$, a state $\sigma$, and a compound action $a$, let
\begin{equation}\label{def:h-o}
\begin{array}{rcl}
h(\sigma,t) & = & \{h(l,t) \,\,|\,\, l \in \sigma \} \\
o(a,t) & = & \{o(a',t) \,\,|\,\, a' \in a \}\mbox{.}
\end{array}
\end{equation}
These sets can be viewed as the representation of $\sigma$ and $a$ in A-Prolog.
%

\begin{definition}
Let $SD$ be an action description of \AL, $n$ be a positive integer, and
$\Sigma(SD)$ be the signature of $SD$. $\Sigma^n_d(SD)$ denotes the signature
obtained as follows:
\begin{itemize}
\item
$const(\Sigma^n_d(SD))= \tbeg  const(\Sigma(SD)) \cup \{ 0, \ldots, n \} \tend$;
\item
$pred(\Sigma^n_d(SD))=\{ h,o \}$.
\end{itemize}
Let
\begin{equation}\label{prg:alpha-d}
\alpha^n_d(SD)=\tbeg \Pi^\alpha_d(SD),\Sigma_d^n(SD) \tend,
\end{equation}
where
\begin{equation}\label{prg:pi-alpha-d}
\Pi^\alpha_d(SD)=\bigcup_{r \in SD} \alpha_d(r),
\end{equation}
and $\alpha_d(r)$ is defined as follows:
\begin{itemize}
\item
$\alpha_d(causes(a,l_0,[l_1,\ldots,l_m]))$ is
\[
h(l_0,T') \hif h(l_1,T),\ldots, h(l_m,T), o(a,T)\mbox{.}
\]
\item
$\alpha_d(caused(l_0,[l_1,\ldots,l_m]))$ is
\begin{equation}\label{map:alpha-d}
h(l_0,T) \hif h(l_1,T), \ldots, h(l_m,T)\mbox{.}
\end{equation}
\item
$\alpha_d(impossible\_if(a,[l_1,\ldots,l_m]))$ is
\[
\begin{array}{rcl}
        & \hif  & h(l_1,T), \ldots, h(l_m,T), \\
        &       & o(a,T)\mbox{.}
\end{array}
\]
\end{itemize}

\st
Let also 
\begin{equation}\label{prg:beta-d}
\beta^n_d(SD)=\tbeg \Pi^\beta_d(SD), \Sigma_d^n(SD) \tend,
\end{equation}
where
\begin{equation}\label{prg:pi-beta-d}
\Pi^\beta_d(SD)=\Pi^\alpha_d(SD) \cup \Pi_d
\end{equation}
and $\Pi_d$ is the following set of rules:
\[
\begin{array}{rlcl}
1\mbox{.}      & h(L,T')       & \hif  & h(L,T), \\
        &               &       & \lpnot h(\overline{L},T')\mbox{.} \\
2\mbox{.}      &               & \hif  & h(L,T), h(\overline{L},T)\mbox{.}
\end{array}
\]

\st
When we refer to a single action description, we will often drop the argument
from $\Sigma_d^n(SD), \alpha^n_d(SD), \Pi^\alpha_d(SD), \beta^n_d(SD),
\Pi^\beta_d(SD)$ in order to simplify the presentation.
\end{definition}

\st
For the rest of this section, we will restrict attention to ground programs. In
order to keep notation simple, we will use $\alpha^n_d$, $\beta^n_d$, $\alpha^n$
and $\beta^n$ to denote the ground versions of the programs previously defined.

\st
For any action description $SD$, state $\sigma_0$ and action
$a_0$, let $\beta^n_d(SD,\sigma_0,a_0)$ denote
\begin{equation}\label{prg:beta-d-sigma-a}
\beta^n_d \cup h(\sigma_0,0) \cup o(a_0,0)\mbox{.}
\end{equation}
We will sometimes drop the first argument, and denote the program by
$\beta^n_d(\sigma_0,a_0)$.

\st
The following lemma will be helpful in proving the main result of this
subsection. It states the correspondence between (single) transitions of the
transition diagram and answer sets of the corresponding A-Prolog program.

\begin{lemma}\label{lemma:Bdprop}
Let $SD$ be an action description, $\mathcal{T}(SD)$ be the transition diagram it
describes, and $\beta^1_d(\sigma_0,a_0)$ be defined as in
(\ref{prg:beta-d-sigma-a}). Then, $\tbeg \sigma_0, a_0, \sigma_1 \tend \in
\mathcal{T}(SD)$ iff $\sigma_1=\{l \,\,|\,\, h(l,1) \in A\}$ for some answer set
$A$ of $\beta^1_d(\sigma_0,a_0)$. 

%

\Begproof
Let us define
\begin{equation}\label{eq:Bdprop:0}
I=h(\sigma_0,0) \cup o(a_0,0)
\end{equation}
and
\[
\beta_d^1 = \beta_d^1(\sigma_0,a_0) \cup I.
\]

\st
\emph{Left-to-right}. Let us show that, if $\tbeg \sigma_0, a_0, \sigma_1 \tend \in
\mathcal{T}(SD)$, 
\begin{equation}\label{eq:Bdprop:constr}
A=I \cup h(\sigma_1,1)
\end{equation}
is an answer set of $\beta^1_d(\sigma_0,a_0)$. Notice that $\tbeg \sigma_0, a_0,
\sigma_1 \tend \in \mathcal{T}(SD)$ implies that $\sigma_1$ is a state.

\st
Let us prove that $A$ is the minimal set of atoms closed under the rules of the
reduct $P^A$. $P^A$ contains:
\begin{enumerate}
\item[a)]
set $I$;
\item[b)]
all rules in $\alpha^1_d(SD)$ (see (\ref{prg:alpha-d}));
\item[c)]
a constraint $\hif h(l,t), h(\overline{l},t)\mbox{.}$ for any fluent literal $l$ and
time point $t$;
\item[d)]
a rule 
\[
h(l,1) \hif h(l,0)
\]
for every fluent literal $l$ such that $h(l,1) \in A$ (in fact, since $\sigma_1$
is complete and consistent, $h(l,1) \in A \iff h(\overline{l},1) \not\in A$).
\end{enumerate}

\st
\underline{$A$ is closed under $P^A$}. We will prove it for every rule of
the program.

\st
Rules of groups (a) and (d): obvious.
%

\st
Rules of group (b) encoding dynamic laws of the form
$causes(a,l,[l_1,\ldots,l_m])$:
\[
\begin{array}{rlcl}
        & h(l,1)        & \hif  & h(l_1,0), \ldots, h(l_m,0), \\
        &               &       & o(a,0)\mbox{.}
\end{array}
\]
If $\{h(l_1,0), \ldots, h(l_m,0), o(a,0)\} \subseteq A$, then, by
(\ref{eq:Bdprop:constr}), $\{l_1,\ldots,l_m\} \subseteq \sigma_0$ and $a \in
a_0$. Therefore, the preconditions of the dynamic law are satisfied by
$\sigma_0$. Hence (\ref{eq:mccain-turner}) implies $l \in \sigma_1$. By
(\ref{eq:Bdprop:constr}), $h(l,1) \in A$.

\st
Rules of group (b) encoding static laws of the form
$caused(l,[l_1,\ldots,l_m])$:
\[
\begin{array}{rlcl}
        & h(l,t)        & \hif  & h(l_1,t), \ldots, h(l_m,t)\mbox{.}
\end{array}
\]
If $\{h(l_1,t), \ldots, h(l_m,t)\} \subseteq A$, then, by
(\ref{eq:Bdprop:constr}), $\{l_1,\ldots,l_m\} \subseteq \sigma_t$, i.e. the
preconditions of the static law are satisfied by $\sigma_t$. If $t=1$, then
(\ref{eq:mccain-turner}) implies $l \in \sigma_1$. By (\ref{eq:Bdprop:constr}),
$h(l,t) \in A$. If $t=0$, since states are closed under the static laws of $SD$,
we have that $l \in \sigma_0$. Again by (\ref{eq:Bdprop:constr}), $h(l,t) \in
A$.

\st
Rules of group (b) encoding impossibility laws of the form
$impossible\_if(a,[l_1, \ldots, l_m])$:
\[
\begin{array}{rlcl}
        &               & \hif  & h(l_1,0), \ldots, h(l_m,0), \\
        &               &       & o(a,0)\mbox{.}
\end{array}
\]
Since $\tbeg \sigma_0, a_0, \sigma_1 \tend \in \mathcal{T}(SD)$ by hypothesis,
$\tbeg \sigma_0, a_0 \tend$ does not satisfy the preconditions of any
impossibility condition. Then, either $a \not\in a_0$ or $l_i \not\in \sigma_0$
for some $i$. By (\ref{eq:Bdprop:constr}), the body of this rule is not
satisfied.

\st
Rules of group (c).
Since $\sigma_0$ and $\sigma_1$ are consistent by hypothesis, $l$ and
$\overline{l}$ cannot both belong to the same state. By (\ref{eq:Bdprop:constr}),
either $h(l,0) \not\in A$ or $h(\overline{l},0) \not\in A$, and the same holds
for time point $1$. Therefore, the body of these rules is never satisfied.

\st
\underline{$A$ is the minimal set closed under the rules of $P^A$}. We will
prove this by assuming that there exists a set $B \subseteq A$ such that $B$ is 
closed under the rules of $P^A$, and by showing that $B=A$.

\st
First of all,
\begin{equation}\label{eq:Bdprop:1}
I \subseteq B,
\end{equation}
since these are facts in $P^A$.

\st
Let 
\begin{equation}\label{eq:Bdprop:2}
\delta = \{l \,\,|\,\, h(l,1) \in B\}\mbox{.}
\end{equation}
Since $B \subseteq A$,
\begin{equation}\label{eq:Bdprop:3}
\delta \subseteq \sigma_1
\end{equation}
We will show that $\delta = \sigma_1$ by proving that
\begin{equation}\label{eq:Bdprop:3-1}
\delta = CN_Z(E(a_0,\sigma_0) \cup (\sigma_1 \cap \sigma_0))\mbox{.}
\end{equation}

\st
\underline{Dynamic laws}. Let $d$ be a dynamic law of $SD$ of the form
$causes(a,l_0,[l_1,\ldots,l_m])$, such that $a \in a_0$ and $\{l_1,\ldots,l_m\}
\subseteq \sigma_0$. Because of (\ref{eq:Bdprop:1}), $h(\{l_1,\ldots,l_m\},0)
\subseteq B$ and $o(a,0) \in B$. Since $B$ is closed under $\alpha_d(d)$
(\ref{map:alpha-d}), $h(l_0,1) \in B$, and $l_0 \in \delta$. Therefore,
$E(a_0,\sigma_0) \subseteq \delta$.

\st
\underline{Inertia}. $P^A$ contains a (reduced) inertia rule of the form
\begin{equation}\label{eq:Bdprop:4}
h(l,1) \hif h(l,0)\mbox{.}
\end{equation}
for every literal $l \in \sigma_1$. Suppose $l \in \sigma_1 \cap \sigma_0$.
Then, $h(l,0) \in I$, and, since $B$ is closed under (\ref{eq:Bdprop:4}), $h(l,1)
\in B$. Therefore, $\sigma_1 \cap \sigma_0 \subseteq \delta$.

\st
\underline{Static laws}. Let $s$ be a static law of $SD$ of the form
$caused(l_0,[l_1,\ldots,l_m])$, such that
\begin{equation}\label{eq:Bdprop:5}
h(\{l_1,\ldots,l_m\},0) \subseteq B\mbox{.}
\end{equation}
Since $B$ is closed under $\alpha_d(s)$ (\ref{map:alpha-d}), $h(l_0,1) \in B$,
and $l_0 \in \delta$. Then, $\delta$ is closed under the static laws of $SD$.

\st
Summing up, (\ref{eq:Bdprop:3-1}) holds. From (\ref{eq:mccain-turner}) and
(\ref{eq:Bdprop:3}), we obtain $\sigma_1 = \delta$. Therefore $h(\sigma_1,1)
\subseteq B$.

\st
At this point we have shown that $I \cup h(\sigma_1,1) \subseteq B \subseteq A$.

\st
\emph{Right-to-left}. Let $A$ be an answer set of $P$ such that $\sigma_1=\{l \,\,|\,\,
h(l,1) \in A \}$. We have to show that
\begin{equation}\label{eq:Bdprop:10}
\sigma_1=CN_Z(E(a_0,\sigma_0) \cup (\sigma_1 \cap \sigma_0)),
\end{equation}
that $\tbeg \sigma_0, a_0 \tend$ respects all impossibility conditions, and that
$\sigma_1$ is consistent and complete.

\st
\underline{$\sigma_1$ consistent}. Obvious, since $A$ is a (consistent) answer set
by hypothesis.

\st
\underline{$\sigma_1$ complete}. By contradiction, let $l$ be a literal s.t. $l
\not\in \sigma_1$, $\overline{l} \not\in \sigma_1$, and $l \in \sigma_0$ (since
$\sigma_0$ is complete by hypothesis, if $l \not\in \sigma_0$, we can still
select $\overline{l}$). Then, the reduct $P^A$ contains a rule
\begin{equation}\label{eq:Bdprop:11}
h(l,1) \hif h(l,0)\mbox{.}
\end{equation}
Since $A$ is closed under $P^A$, $h(l,1) \in A$ and $l \in \sigma_1$.
Contradiction.

\st
\underline{Impossibility conditions respected}. By contradiction, assume that
condition $\\impossible\_if(a,[l_1,\ldots,l_m])$ is not respected. Then,
$h(\{l_1,\ldots,l_m\},0) \subseteq A$ and $\\o(a,0) \in A$. Therefore, the body of
the $\alpha_d$-mapping (\ref{map:alpha-d}) of the impossibility condition is
satisfied by $A$, and $A$ is not a (consistent) answer set.

\st
\underline{(\ref{eq:Bdprop:10}) holds}. Let us prove that $\sigma_1 \supseteq
E(a_0,\sigma_0)$. Consider a dynamic law $d$ in $SD$ of the form
$causes(a,l_0,[l_1, \ldots, l_m])$, such that $\{ l_1, \ldots, l_m \} \subseteq
\sigma_0$ and $a \in a_0$. Since $A$ is closed under $\alpha_d(d)$
(\ref{map:alpha-d}), $h(l_0,1) \in A$. Then, $\sigma_1 \supseteq
E(a_0,\sigma_0)$. 

\st
$\sigma_1 \supseteq \sigma_1 \cap \sigma_0$ is trivially true.

\st
Let us prove that $\sigma_1$ is closed under the static laws of $SD$. Consider a
static law $s$, of the form $caused(l_0,[l_1, \ldots, l_m])$, such that $\{l_1,
\ldots, l_m\} \subseteq \sigma_0$. Since $A$ is closed under $\alpha_d(s)$
(\ref{map:alpha-d}), $h(l_0,1) \in A$.

\st
Let us prove that $\sigma_1$ is the minimal set satisfying all conditions.
By contradiction, assume that there exists a set $\delta \subset \sigma_1$
such that $\delta \supseteq E(a_0,\sigma_0) \cup (\sigma_1 \cap \sigma_0)$ and
that $\delta$ is closed under the static laws of $SD$. We will prove
that this implies that $A$ is not an answer set of $P$.

\st
Let $A'$ be the set obtained by removing from $A$ all atoms $h(l,1)$ such that
$l \in \sigma_1 \setminus \delta$. Since $\delta \subset \sigma_1$, $A' \subset
A$.

\st
Since $\delta \supseteq E(a_0,\sigma_0) \cup (\sigma_1 \cap
\sigma_0)$, for every $l \in \sigma_1 \setminus \delta$ it must be
true that $l \not\in \sigma_0$ and $l \not\in E(a_0,\sigma_0)$.
Therefore there must exist (at least) one static law
$caused(l,[l_1,\ldots,l_m])$ such that $\{l_1, \ldots, l_m \} \subseteq
\sigma_1$ and $\{l_1, \ldots, l_m \} \not\subseteq \delta$. Hence, $A'$ is
closed under the rules of $P^A$. This proves that $A$ is not an answer set of
$P$. Contradiction.
%
%
\Endproof
\end{lemma}

\st
We are now ready to prove the main result of this subsection. The following
notation will be used in the theorems that follow. Let $SD$ be an 
action description, and $M=\tbeg \sigma_0, a_0, \sigma_1, \ldots, a_{n-1},
\sigma_n \tend$ be a sequence where $\sigma_i$ are sets of fluent literals and
$a_i$ are actions. $o(M)$ denotes
\[
\bigcup_t \,o(a_t,t),
\]
with $o(a,t)$ from (\ref{def:h-o}). The length of $M$, denoted by $l(M)$, is
$\frac{m-1}{2}$, where $m$ is the number of elements of $M$.

\st
Given an action description $SD$ and a sequence $M=\tbeg \sigma_0, a_0,
\sigma_1, \ldots, a_{n-1},\sigma_n \tend$, $\beta^n_d(SD,M)$ denotes the 
program
\begin{equation}\label{prg:beta-d-M}
\beta_d^n \cup h(\sigma_0,0) \cup o(M)\mbox{.}
\end{equation}
We will use $\beta^n_d(M)$ as short form for $\beta^n_d(SD,M)$.

\begin{lemma}\label{lemma:premain}
Let $SD$ be an action description, $M=\tbeg \sigma_0, a_0,\sigma_1, \ldots,
a_{n-1},\sigma_n \tend$ be a sequence of length $n$, and $\beta^n_d(M)$ be
defined as in (\ref{prg:beta-d-M}). If $\sigma_0$ is a state, then, $M$ is a
trajectory of $\mathcal{T}(SD)$ iff $M$ is defined by an answer set of
$P=\beta^n_d(M)$.
\if T\extPaper
\Begproof
By induction on $l(M)$.

\st
Base case: $l(M)=1$. Since $l(M)=1$, $M$ is the sequence $\tbeg \sigma_0,
a_0, \sigma_1 \tend$, $o(M)=o(a_0,0)$ and, since $\sigma_0$ is a state by
hypothesis, $P=\beta^1_d(SD,\sigma_0,a_0)$ (\ref{prg:beta-d-sigma-a}). Then,
Lemma \ref{lemma:Bdprop} can be applied, thus completing the proof for this
case.

\st
Inductive step. We assume that the theorem holds for trajectories of length $n-1$
and prove that it holds for trajectories of length $n$.

\st
$M=\tbeg \sigma_0, a_0, \sigma_1, \ldots, a_{n-1}, \sigma_n \tend$ is a
trajectory of $\mathcal{T}(SD)$ iff
\begin{eqnarray}
& \tbeg \sigma_0, a_0, \sigma_1 \tend \in \mathcal{T}(SD) \mbox{, and } &
\label{eq:premain:1}\\
& M'=\tbeg \sigma_1, a_1, \ldots, a_{n-1},\sigma_n \tend \mbox{ is a trajectory
of } \mathcal{T}(SD)\mbox{.} & \label{eq:premain:2}
\end{eqnarray}

\st
Let $R_1$ denote $\beta^1_d(SD,\sigma_0,a_0)$. By Lemma \ref{lemma:Bdprop},
(\ref{eq:premain:1}) holds iff 
\begin{equation}\label{eq:premain:3}
R_1 \mbox{ has an answer set, } A \mbox{, such that } \sigma_1=\{l \,\,|\,\,
h(l,1) \in A\}\mbox{.}
\end{equation}

\st
Notice that this implies that $\sigma_1$ is a state. Now, let $R_2$ denote
$\beta^{n-1}_d(SD,M')$. By inductive hypothesis, (\ref{eq:premain:2}) holds iff
\begin{equation}\label{eq:premain:4}
M' \mbox{ is defined by an answer set, } B \mbox{, of } R_2\mbox{.}
\end{equation}

\st
Let $S$ be the set containing all literals of the form $h(l,0)$, $h(l,1)$ and
$o(a,0)$, over the signature of $P$. Let $C$ be the set of the constraints
of $R_1$. Notice that:
\begin{itemize}
\item
$S$ is a splitting set of $P$;
\item
$bottom_S(P)=R_1 \setminus C$.
\end{itemize}
Then, (\ref{eq:premain:3}) holds iff
\begin{eqnarray}
& A \mbox{ is an answer set of } bottom_S(P), \mbox{ satisfying } C, & \nonumber\\
& \mbox{such that } \sigma_1=\{ l \,\,|\,\, h(l,1) \in A \}\mbox{.} &
\label{eq:premain:5}
\end{eqnarray}

\st
For any program $R$, let $R^{+1}$ denote the program obtained from $R$ by:
\begin{enumerate}
\item
replacing, in the rules of $R$, every occurrence of a constant symbol denoting a
time point with the constant symbol denoting the next time point;
\item
modifying accordingly the signature of $R$.
\end{enumerate}

\st
Then, (\ref{eq:premain:4}) holds iff
\begin{equation}\label{eq:premain:6}
M' \mbox{ is defined by a set } B \mbox{ such that } B^{+1} \mbox{ is an answer set
of } R_2^{+1}.
\end{equation}

\st
Notice that $e_S(P,A) = e_S(R_2^{+1},A) \cup e_S(C,A)$. Therefore, $A$ satisfies
$C$ iff $e_S(P,A)=e_S(R_2^{+1},A)$. Notice also that $e_S(R_2^{+1},A) \simeq
R_2^{+1}$.

\st
Then, (\ref{eq:premain:5}) and (\ref{eq:premain:6}) hold iff
\begin{eqnarray}
& A \mbox{ is an answer set of } bottom_S(P) & \nonumber \\
& \mbox{such that } \sigma_1=\{l \,\,|\,\, h(l,1) \in A \} \mbox{, and} &
\nonumber \\
& M' \mbox{ is defined by } B \mbox{, and} & \label{eq:premain:7} \\
& B^{+1} \mbox{ is an answer set of } e_S(P,A)\mbox{.} & \nonumber
\end{eqnarray}

\st
By Definition \ref{def:defines}, (\ref{eq:premain:7}) holds iff
\begin{eqnarray}
& A \mbox{ is an answer set of } bottom_S(P) \mbox{, and} & \nonumber \\
& M \mbox{ is defined by } A \cup B^{+1} \mbox{, and} & \label{eq:premain:8} \\
& B^{+1} \mbox{ is an answer set of } e_S(P,A)\mbox{.} & \nonumber
\end{eqnarray}

\st
By the Splitting Set Theorem, (\ref{eq:premain:8}) holds iff
\[
M \mbox{ is defined by an answer set of } P.
\]
\Endproof
\fi
\end{lemma}

\subsubsection{Step 2}
In this subsection, we extend the previous encoding in order to be able to
generate programs whose answer sets describe exactly the paths consistent with a
specified recorded history $\Gamma_n$.

%

\st
Let $\Sigma^n_{\Gamma,d}(SD)$ denote the signature defined as
follows:
\begin{itemize}
\item
$const(\Sigma^n_{\Gamma,d}(SD))=const(\Sigma^n_d(SD))$;
\item
$pred(\Sigma^n_{\Gamma,d}(SD))=pred(\Sigma^n_d(SD)) \cup \{ hpd,obs \}$.
\end{itemize}
Let 
\begin{equation}\label{prg:alpha-sd-d}
\alpha^n_d(SD,\Gamma_n)=\tbeg \Pi^\Gamma_d,\Sigma^n_{\Gamma,d} \tend,
\end{equation}
where
\begin{equation}\label{prg:pi-gamma-d}
\Pi^\Gamma_d = \Pi^\beta_d(SD) \cup \hat{\Pi} \cup \Gamma_n\mbox{.}
\end{equation}
$\Pi^\beta_d(SD)$ is defined as in (\ref{prg:pi-beta-d}), and $\hat{\Pi}$ is the set
of rules:
\[
\begin{array}{rlcl}
3\mbox{.}      & o(A,T)        & \hif  & hpd(A,T)\mbox{.} \\
4\mbox{.}      & h(L,0)        & \hif  & obs(L,0)\mbox{.} \\
5\mbox{.}     &               & \hif  & obs(L,T), \\
        &               &       & \lpnot h(L,T)\mbox{.}
\end{array}
\]
(Notice that these rules are equal the last $3$ rules of program $\Pi$,
defined in Section \ref{sec:computing}.)

\st
When we refer to a single system description, we will often drop argument $SD$
from 
$\Sigma^n_{\Gamma,d}(SD), \alpha_d^n(SD,\Gamma_n), \Pi^\Gamma_d(SD)$ in order to
simplify the presentation.

\st
Notice that, as we did before, in the rest of this section we will restrict
attention to ground programs.

\begin{proposition}\label{proposition:dmain}
If the initial situation of $\Gamma_n$ is complete, i.e. for any fluent $f$ of
$SD$, $\Gamma_n$ contains $obs(f,0)$ or $obs(\neg f,0)$, then 
\begin{equation}\label{th:dmain:1}
M \mbox{ is a model of } \Gamma_n
\end{equation}
iff
\begin{equation}\label{th:dmain:2}
M \mbox{ is defined by some answer set of } \alpha^n_d(SD,\Gamma_n)\mbox{.}
\end{equation}
\if T\extPaper
\Begproof
Let $P_0$ be $\alpha^n_d(SD,\Gamma_n)$, and $P_1$ be obtained
from $P_0$ by removing every constraint
\begin{equation}\label{eq:dmain:1}
\begin{array}{cl}
\hif    & obs(l,t), \\
        & \lpnot h(l,t)\mbox{.}
\end{array}
\end{equation}
such that $obs(l,t) \not\in \Gamma_n$. Notice that
\begin{equation}\label{eq:dmain:2}
P_0 \simeq P_1,
\end{equation}
hence (\ref{th:dmain:2}) holds iff
\begin{equation}\label{eq:dmain:3}
M \mbox{ is defined by some answer set of } P_1\mbox{.}
\end{equation}

\st
Let $Q$ be the set of literals, over the signature of $P_1$, of the form
$hpd(A,T)$ and $obs(L,0)$. Let $\vec{q}$ be an arbitrary ordering of
the elements of $Q$, and $P_2$ be $t(P_1,\vec{q})$. By Lemma
\ref{lemma:strongext}, $P_1 \succ_Q P_2$. Since predicate names $h$ and
$o$ are common to the signatures of both $P_1$ and $P_2$, (\ref{eq:dmain:3})
holds iff
\begin{equation}\label{eq:dmain:4}
M \mbox{ is defined by some answer set of } P_2\mbox{.}
\end{equation}

\st
Let $\sigma_0$ be $\{ l \,\,|\,\, obs(l,0) \in \Gamma_n \}$. It is easy to check
that
\[
rules(P_2)=R \cup C,
\]
where $R=\beta_d^n \cup h(\sigma_0,0) \cup \{ o(a,t) \,\,|\,\, hpd(a,t) \in
\Gamma_n \}$ (see (\ref{prg:pi-beta-d})) and $C$ is the set of constraints
\[
\hif \lpnot h(l,t)\mbox{.}
\]
such that $obs(l,t) \in \Gamma_n$.

\st
Then, (\ref{eq:dmain:4}) holds iff
\begin{eqnarray}
& M \mbox{ is defined by some answer set}, A, \mbox{ of } R, \mbox{and} &
\label{eq:dmain:5}\\
& A \mbox{ satisfies } C\mbox{.} & \label{eq:dmain:6}
\end{eqnarray}

\st
Notice that $R=\beta_d^n(M)$. 

Since the initial situation of $\Gamma_n$ is complete, $\sigma_0$ is complete.
This, together with (\ref{eq:dmain:5}), implies that $\sigma_0$ is closed under
the static laws of $SD$, i.e. $\sigma_0$ is a state. Then, Lemma
\ref{lemma:premain} can be applied, obtaining that (\ref{eq:dmain:5}) holds iff 
\begin{eqnarray}
& M \mbox{ is a trajectory of } \mathcal{T}(SD), \mbox{and} & \nonumber \\
& o(M)=\{o(a,t) \,\,|\,\, hpd(a,t) \in \Gamma_n \} & \label{eq:dmain:7}
\end{eqnarray}

\st
By construction of $C$, (\ref{eq:dmain:6}) holds iff, if $obs(l,t) \in
\Gamma_n$, then $h(l,t) \in A$. According to Lemma \ref{lemma:premain}, $h(l,t)
\in A$ iff $l \in \sigma_t$, where $\sigma_t$ are the states that appear in $M$.
Therefore, (\ref{eq:dmain:6}) holds iff
\begin{equation}\label{eq:dmain:8}
\mbox{if } obs(l,t) \in \Gamma_n, \mbox{then } l \in \sigma_t\mbox{.}
\end{equation}

\st
By Definition \ref{model}(a), (\ref{eq:dmain:7}) and (\ref{eq:dmain:8}) hold
iff
\[
M \mbox{ is a model of } \Gamma_n\mbox{.}
\]
\Endproof
\fi
\end{proposition}

\subsubsection{Step 3}
In this section we prove that models of $\Gamma_n$ are in a one-to-one
correspondence with the answer sets of $\alpha(SD,\Gamma_n)$. We will do this by
proving that the answer sets of $\alpha_d^n(SD,\Gamma_n)$ and of
$\alpha(SD,\Gamma_n)$ define the same models.

\st
In order to prove this equivalence, we define a new encoding of \AL that
will allow us to link $\alpha(SD,\Gamma_n)$ and $\alpha_d^n(SD,\Gamma_n)$.

\st
Let $SD$ be an action description of \AL and $\Sigma(SD)$ be its signature. For
any positive integer $n$, $\Sigma^n(SD)$ denotes the signature obtained as
follows:
\begin{itemize}
\item
$const(\Sigma^n(SD))=const(\Sigma(SD)) \cup \{0, \ldots, n\} \cup
\{1, \ldots, k\}$, where $k$ is the maximum number of preconditions present in
the laws of $SD$;
\item
$pred(\Sigma^n(SD))=\{h,o,d\_law,s\_law,head,action,prec,all\_h,prec\_h\}$.
\end{itemize}

\st
Let
\begin{equation}\label{prg:alpha}
\alpha^n(SD)=\tbeg \Pi^\alpha(SD), \Sigma^n(SD) \tend,
\end{equation}
where
\begin{equation}\label{prg:pi-alpha}
\Pi^\alpha(SD)=\bigcup_{r \in SD} \alpha(r)
\end{equation}
and $\alpha(r)$ is defined as in Section \ref{sec:computing}.

\st
Finally, let
\begin{equation}\label{prg:beta}
\beta^n(SD)=\tbeg \Pi^\beta(SD), \Sigma^n(SD) \tend,
\end{equation}
where
\begin{equation}\label{prg:pi-beta}
\Pi^\beta(SD) = \Pi^\alpha(SD) \cup \Pi^{-}
\end{equation}
and $\Pi^{-}$ is the set of rules:
\[
\begin{array}{rlcl}
1\mbox{.} &h(L,T^{\prime}) & \leftarrow & d\_law(D),\\
          &           &            & head(D,L),\\
          &           &            & action(D,A),\\
          &           &            & o(A,T),\\
          &           &            & prec\_h(D,T)\mbox{.}\\
2\mbox{.} &h(L,T)          & \leftarrow & s\_law(D),\\
          &           &            & head(D,L),\\
          &           &            & prec\_h(D,T)\mbox{.}\\ 
3\mbox{.} &all\_h(D,N,T)   & \leftarrow & prec(D,N,nil)\mbox{.} \\
4\mbox{.} &all\_h(D,N,T)   & \leftarrow & prec(D,N,P), \\
          &           &            & h(P,T), \\
          &           &            & all\_h(D,N^\prime,T)\mbox{.} \\
5\mbox{.} &prec\_h(D,T)    & \leftarrow & all\_h(D,1,T)\mbox{.}\\
6\mbox{.} &h(L,T^\prime)   & \leftarrow & h(L,T),\\
          &           &            & \no h(\overline{L}, T^{\prime})\mbox{.}\\
7\mbox{.} &           &        \leftarrow & h(L,T), h(\overline{L},T). \\
%
%
%
%
%
\end{array}
\]
(Notice that these rules correspond to rules $(1)-(7)$ of program $\Pi$ defined
in Section \ref{sec:computing}.)

\st
When we refer to a single action description, we will often drop the argument
from $\Sigma^n(SD), \alpha^n(SD), \Pi^\alpha(SD), \beta^n(SD), \Pi^\beta(SD)$ in
order to simplify the presentation. We will also restrict attention to the
ground versions of the programs just defined. For this reason, we will abuse
notation slightly and denote by $\alpha^n(SD)$ and $\beta^n(SD)$ the 
ground versions of the programs defined above.

\st
The following theorem establishes a link between $\beta^n$ and $\beta_d^n$.

\begin{lemma}\label{lemma:simpl}
Let $SD$ be an action description, $n$ be a positive integer, and $Q$ denote
$lit(\beta^n) \setminus lit(\beta^n_d)$. Then, for any program $R$
such that $lit(R) \cap Q = \emptyset$,
\[
\beta^n \cup R \succ_Q \beta^n_d \cup R\mbox{.}
\]

\Begproof
Let $\vec{q}$ be an ordering of the elements of $Q$, $P$ be $\beta^n \cup
R$, and $P_d$ be $\beta^n_d \cup R$.

\st
Notice that, in $e(P,\vec{q})$, the elements of $Q$ only occur in the rules that
define them, and that $lit(R) \cap Q=\emptyset$ by hypothesis. Then, by Lemma
\ref{lemma:strongext},
\begin{equation}\label{eq:simpl:1}
P \succ_Q t(P,\vec{q})\mbox{.}
\end{equation}
It can also be easily checked that $t(P,\vec{q})=P_d$. Hence, (\ref{eq:simpl:1})
can be rewritten as
\[
P \succ_Q P_d,
\]
that is,
\[
\beta^n \cup R \succ_Q \beta^n_d \cup R\mbox{.}
\]
\Endproof
\end{lemma}

\st
We are finally able to give the proof of Theorem \ref{th1}.

\st
\emph{Theorem \ref{th1}} \\
If the initial situation of $\Gamma_n$ is complete, i.e. for any fluent $f$ of
$SD$, $\Gamma_n$ contains $obs(f,0)$ or $obs(\neg f,0)$, then $M$ is a model of
$\Gamma_n$ iff $M$ is defined by some answer set of $\alpha(SD,\Gamma_n)$.

\Begproof
By Proposition \ref{proposition:dmain}, $M$ is a model of $\Gamma_n$ iff
$M$ is defined by some answer set of $\alpha^n_d(SD,\Gamma_n)$. Let $P_d$ be
$\alpha^n_d(SD,\Gamma_n)$ and $P$ be $\alpha(SD,\Gamma_n)$.
Let $R$ be $P \setminus \beta^n$. Let also $Q$ be $lit(P) \setminus
lit(P_d)$. By Lemma \ref{lemma:simpl}, $\beta^n \cup R \succ_Q \beta_d^n \cup
R$.
From this we obtain that 
$\alpha(SD,\Gamma_n) \succ_Q \alpha^n_d(SD,\Gamma_n)$. Notice that predicate
names $h$ and $o$ are common to the signatures of both $P$ and $P_d$. Then, the
thesis follows from the definition of Strong Conservative Extension.
\Endproof

\st
The following corollary extends Theorem \ref{th1} to the case in which the
initial situation of $\Gamma_n$ is not complete.

\begin{corollary}\label{corollary:main_incomp}
Let $R$ be
\[
\begin{array}{rcl}
h(F,0)        & \hif  & \lpnot h(\neg F,0)\mbox{.} \\
h(\neg F,0)   & \hif  & \lpnot h(F,0)\mbox{.}
\end{array}
\]
For any history $\Gamma_n$,
\begin{equation}\label{th:main_incomp-1}
M \mbox{ is a model of } \Gamma_n
\end{equation}
iff
\begin{equation}\label{th:main_incomp-2}
M \mbox{ is defined by some answer set of } \alpha(SD,\Gamma_n) \cup R\mbox{.}
\end{equation}

\Begproof
%
Let $\sigma_0$ be the first component of $M$, and $obs(\sigma_0,0)=\{obs(l,0) \,\,|\,\, l \in \sigma_0\}$.
First of all, we will show that (\ref{th:main_incomp-2}) is equivalent to
\begin{equation}\label{eq:main_incomp-0-1}
M \mbox{ is defined by some answer set of } \alpha(SD,\Gamma_n) \cup
obs(\sigma_0,0)\mbox{.}
\end{equation}
Let $\Pi_1 = \alpha(SD,\Gamma_n) \cup R$ and $\Pi_2 = \alpha(SD,\Gamma_n) \cup obs(\sigma_0,0)$.
Consider a splitting of $\Pi_1$ and $\Pi_2$ based on set
\[
S = \{ obs(l,0) \,\,|\,\, l \in FL \} \cup \{ h(l,0) \,\,|\,\, l \in FL \}
\]
where $FL$ is the set of fluent literals of $SD$. Let set $Q$ include:
\begin{itemize}
\item
the set of ground instances, with $T = 0$, of rules (2)-(5) (we need to consider only the
instances where variable $D$ denotes a static law), and (9) from program $\Pi$, in Section
\ref{sec:computing};
\item
the subset of $\alpha(SD)$ containing those facts occurring in the body of the above rules.
\end{itemize}
$\Pi_1$ and $\Pi_2$ are split by $S$ so that:
\begin{eqnarray}
&&bottom_S(\Pi_1) = R \cup Q \cup (\Gamma_n \cap S)\mbox{,}\\
&&bottom_S(\Pi_2) = Q \cup obs(\sigma_0,0)\mbox{,}\\
&&top_S(\Pi_1) = top_S(\Pi_2) \label{eq:main_incomp-0-2}\mbox{.}
\end{eqnarray}
$bottom_S(\Pi_2)$ has a unique answer set, $A_2$, and $A_2 \cap lit(h)=h(\sigma_0,0)$. It can be shown
that there exists an answer set, $A_1$, of $bottom_S(\Pi_1)$, such that $A_1 \subseteq A_2$. 
Moreover, for such $A_1$, 
\begin{equation}\label{eq:main_incomp-0-3}
A_1 \setminus lit(obs) = A_2 \setminus lit(obs)\mbox{.}
\end{equation}
Let $Q'$ denote the set of ground instances, with $T = 0$, of rule (10) from program $\Pi$ in Section 
\ref{sec:computing}. From (\ref{eq:main_incomp-0-2}) and (\ref{eq:main_incomp-0-3}),
\begin{equation}\label{eq:main_incomp-0-4}
e_S(\Pi_1 \setminus Q',A_1) \simeq e_S(\Pi_2 \setminus Q',A_2)\mbox{.}
\end{equation}
Since the body of the rules in $Q'$ is never satisfied,
\begin{equation}\label{eq:main_incomp-0-5}
e_S(\Pi_1,A_1) \simeq e_S(\Pi_2,A_2)\mbox{.}
\end{equation}
Let $B$ be an answer set of $e_S(\Pi_2,A_2)$ and $C_2 = B \cup A_2$. By the Splitting Set Theorem,
$C_1 = B \cup A_1$ is an answer set of $\Pi_1$. This implies that (\ref{th:main_incomp-2}) is
equivalent to (\ref{eq:main_incomp-0-1}).
%
%
%

\st
Now we will complete the proof by showing that (\ref{eq:main_incomp-0-1}) is equivalent
to (\ref{th:main_incomp-1}). Since
\[
\alpha(SD,\Gamma_n) \cup obs(\sigma_0,0) = \alpha(SD,\Gamma_n \cup
obs(\sigma_0,0)),
\]
Equation (\ref{eq:main_incomp-0-1}) holds iff
\begin{equation}\label{eq:main_incomp-3}
M \mbox{ is defined by some answer set of } \alpha(SD,\Gamma_n \cup
obs(\sigma_0,0))\mbox{.}
\end{equation}
By Theorem \ref{th1}, (\ref{eq:main_incomp-3}) holds iff
\begin{equation}\label{eq:main_incomp-4}
M \mbox{ is a model of } \Gamma_n \cup obs(\sigma_0,0)\mbox{.}
\end{equation}
By Definition \ref{model}(a), (\ref{eq:main_incomp-4}) holds iff
\begin{equation}\label{eq:main_incomp-5}
M \mbox{ is a model of } \Gamma_n.
\end{equation}
\Endproof
\end{corollary}

%

\subsection{Answer sets of $D_0(\mathcal{S})$ and candidate diagnoses}
Theorem \ref{theorem:candidate_diag} establishes a link between
answer sets and candidate diagnoses. In this section, we give a proof of this
theorem.

\st
\emph{Theorem \ref{theorem:candidate_diag}}\\
Let $\langle \Sigma, T, W \rangle$ be a diagnostic domain, $SD$ be a system
description of $T$, $\mathcal{S}=\tbeg  \Gamma_n, O^m_n \tend$ be
a symptom of the system's malfunctioning, and $E$ and $\delta$ be sets
of ground atoms.
\begin{equation}\label{th:candidate_diag-1}
\langle E, \Delta \rangle \mbox{ is a candidate diagnosis of } \mathcal{S}
\end{equation}
iff
\begin{equation}\label{th:candidate_diag-2}
\langle E,\Delta \rangle \mbox{ is determined by an answer set of }
D_0(\mathcal{S})\mbox{.}
\end{equation}

\Begproof
By definition of candidate diagnosis, (\ref{th:candidate_diag-1}) holds
iff
\begin{equation}\label{eq:candidate_diag-1}
\begin{array}{c}
\mbox{there exists a model, $M$, of history } \Gamma_n \cup O^m_n \cup E
\mbox{ such that} \\
\Delta=\{ c \,\,|\,\, M \entails h(ab(c),m) \}\mbox{.}
\end{array}
\end{equation}
By Corollary \ref{corollary:main_incomp}, (\ref{eq:candidate_diag-1}) holds iff
\begin{equation}\label{eq:candidate_diag-2}
\begin{array}{c}
\mbox{there exists an answer set, $X$, of }
P=Conf(\mathcal{S}) \cup E
\mbox{ such that} \\
\Delta=\{ c \,\,|\,\, h(ab(c),m) \in X \}\mbox{.}
\end{array}
\end{equation}
Consider now (\ref{th:candidate_diag-2}). By definition \ref{def:D0-diagnosis},
(\ref{th:candidate_diag-2}) holds iff
\begin{equation}\label{eq:candidate_diag-3}
\begin{array}{c}
\mbox{there exists an answer set, $X'$, of $P'=D_0(\mathcal{S})$ such that} \\
\langle E, \Delta \rangle \mbox{ is determined by } X'.
\end{array}
\end{equation}

\st
Let
\[
SP_0=\{ hpd(a,t) \,\,|\,\, hpd(a,t) \in P \land a \in A_e \},
\]
\[
SP=SP_0 \cup \{ o(a,t) \,\,|\,\, hpd(a,t) \in SP_0 \}\mbox{.}
\]
$SP$ is a splitting set for both $P$ and $P'$.

\st
By the Splitting Set Theorem, (\ref{eq:candidate_diag-2}) and
(\ref{eq:candidate_diag-3}) are, respectively, equivalent to
\begin{equation}\label{eq:candidate_diag-2-1}
\begin{array}{c}
\mbox{(a) there exists an answer set, $X_B$, of $bottom_{SP}(P)$ and} \\
\mbox{(b) there exists an answer set, $Y$, of $e_{SP}(P,X_B)$} \\
\mbox{such that } \Delta=\{ c \,\,|\,\, h(ab(c),m) \in X_B \cup Y \}\mbox{.}
\end{array}
\end{equation}
\begin{equation}\label{eq:candidate_diag-3-1}
\begin{array}{c}
\mbox{(a) there exists an answer set, $X'_B$, of $bottom_{SP}(P')$ and}\\
\mbox{(b) there exists an answer set, $Y'$, of $e_{SP}(P',X'_B)$} \\
\mbox{such that } \langle E, \Delta \rangle \mbox{ is determined by } X'_B \cup Y'.
\end{array}
\end{equation}
We want to prove that (\ref{eq:candidate_diag-2-1}) and
(\ref{eq:candidate_diag-3-1}) are equivalent.

\st
Let
\[
E_d=\{ hpd(a,t) \,\,|\,\, hpd(a,t) \in E \land hpd(a,t) \not\in \Gamma_n\},
\]
and $Q$ be the set of rules
\[
o(A,T) \hif hpd(A,T)\mbox{.}
\]
for $A$ and $T$ such that $hpd(A,T) \in E_d$. $bottom_{SP}(P)$ is $E_d \cup Q$
and $bottom_{SP}(P')$ is $DM_0 \cup Q$. Let $B$ and $B'$ be defined as
follows:
\begin{equation}\label{eq:candidate_diag-4}
\begin{array}{c}
H=\{ hpd(a,t) \,\,|\,\, hpd(a,t) \in E_d \} \\
B=H \cup \{ o(a,t) \,\,|\,\, hpd(a,t) \in H \} \\
B'=B \setminus E_d \cup \{ \neg o(a,t) \,\,|\,\, o(a,t) \not\in B \}\mbox{.}
\end{array}
\end{equation}
Let us show that (\ref{eq:candidate_diag-2-1}a) holds iff
(\ref{eq:candidate_diag-3-1}a) holds.

\st
Assume that (\ref{eq:candidate_diag-2-1}a) holds. It is easy to see that $B$ is
the unique answer set of $bottom_{SP}(P)$. If we observe that the answer sets of
$bottom_{SP}(P')$ enumerate all possible sequences of exogenous actions, we
obtain that $B'$ is an answer set of $bottom_{SP}(P')$. Therefore,
(\ref{eq:candidate_diag-3-1}a) holds.

\st
Now, assume that (\ref{eq:candidate_diag-3-1}a) holds. As before, $B'$ is an
answer set of $bottom_{SP}(P')$. Immediately, we obtain that $B$ is an answer
set of $bottom_{SP}(P)$. Therefore, (\ref{eq:candidate_diag-2-1}a) holds.

\st
Let us now show that $e_{SP}(P,B)=e_{SP}(P',B')$. Notice that
$top_{SP}(P)=Conf(\mathcal{S}) \setminus Q=top_{SP}(P')$. Let $I=B \cap B'$ and
$\overline{I}=(B \cup B') \setminus I$. Observe that, for every literal $l \in
\overline{I}$, $l \not\in top_{SP}(P)$. This means that
\begin{equation}\label{eq:candidate_diag-6}
e_{SP}(P,B)=e_{SP}(P,I)=e_{SP}(P',I)=e_{SP}(P',B')\mbox{.}
\end{equation}

\st
Let $Z$ denote an answer set of $e_{SP}(P,B)$. By construction of $B$ and $B'$
and from (\ref{eq:candidate_diag-6}), 
\[
\Delta=\{ c \,\,|\,\, h(ab(c),n-1) \in (B \cup Z)\} \mbox{ iff } \langle E, \Delta \rangle \mbox{ is
determined by } B' \cup Z.
\]
Hence, (\ref{eq:candidate_diag-2-1}) and (\ref{eq:candidate_diag-3-1}) are
equivalent.
\Endproof

\subsection{Properties of Diagnostic Module $D_1$}
In this subsection, we show that using diagnostic module $D_1$
in place of $D_0$ is \emph{safe},
i.e. $D_1$ will not miss any useful
predictions about the malfunctioning components.

\st
We start by introducing some terminology which is needed in the rest of the subsection.
\begin{definition}
Elementary action $a_e$ is \emph{relevant to fluent literal $l$} (written $rel(a_e,l)$), if:
\begin{itemize}
\item
$\mbox{``}causes(a_e,l,P)\mbox{''} \in SD$;
\item
$\mbox{``}caused(l,[l_1,\ldots,l_m])\mbox{''} \in SD$ and $a_e$ is relevant to some $l_i$
from the preconditions of the law;
\item
$\mbox{``}impossible\_if(a_e',[l_1,\ldots,l_m])\mbox{''} \in SD$, $a_e'$ is relevant to some $l$, and
$a_e$ is relevant to some $\overline{l_i}$ from the preconditions of the condition.
\end{itemize}

\st
An action, $a$, is \emph{relevant to set $O$} of fluent literals if every elementary action from $a$
is relevant to some $l \in O$.
\end{definition}
\begin{definition}
The set, $rel(O)$, of fluent literals \emph{relevant to collection of fluent literals $O$} is defined
as follows.
\begin{enumerate}
\item
$O \subseteq rel(O)$;
\item
if $\mbox{``}causes(a_e,l,P)\mbox{''} \in SD$, and $l \in rel(O)$, then $P \subseteq rel(0)$;
\item
if $\mbox{``}caused(l,P)\mbox{''} \in SD$, and $l \in rel(O)$, then $P \subseteq rel(0)$;
\item
for every condition ``$impossible\_if(a_e,[l_1,\ldots,l_m])$'' from $SD$,
if $a_e$ is relevant to $O$, then
$\{ \overline{l_1}, \ldots, \overline{l_m} \} \subseteq rel(O)$.
\end{enumerate}
\end{definition}
\begin{definition}
States $s_1$,$s_2$ are called \emph{equivalent w.r.t. a set of fluent literals $O$}
($s_1 \eqO{O} s_2$) if
\[
\forall l \in rel(O) \,\, l \in s_1 \mbox{ iff } l \in s_2\mbox{.}
\]
\end{definition}
\begin{definition}
The \emph{rank of a sequence of actions, $\alpha$, w.r.t. a collection of fluent
literals, $O$}, is the number of elementary actions in $\alpha$ which are not
relevant to $O$, and is denoted by $|\alpha|_O$.
\end{definition}
\begin{definition}
Sequence of actions $\alpha_2$ is the \emph{reduct of $\alpha_1$ w.r.t. $O$}
($\alpha_2=red_O(\alpha_1)$) if $\alpha_2$ is obtained from $\alpha_1$ by
replacing $a$ by $a \setminus \{ a_e \}$ if $a_e$ is an exogenous action
from $a$ not relevant to $O$. We say that $\alpha_1$ is \emph{equivalent to
$\alpha_2$ w.r.t. $O$} ($\alpha_1 \eqO{O} \alpha_2$) if $\alpha_1=red_O(\alpha_2)$
or $\alpha_2=red_O(\alpha_1)$.
\end{definition}
\begin{definition}[Well-defined system description]
Let $T$ be the transition diagram corresponding to system description $SD$.
$SD$ is \emph{well-defined} if, for any state $s$ and action $a$,
$\mbox{``}impossible\_if(a,P)\mbox{''} \in SD$  and $P \subseteq s$
iff there is no $s'$ such that $\langle s,a,s' \rangle \in T$.
\end{definition}
\begin{definition}[Set of final states]
Let $O$ be a collection of fluent literals. The set of final states w.r.t. $O$ is
\[
F_O = \{ s \,\,|\,\, O \subseteq s \}\mbox{.}
\]
\end{definition}

\begin{definition}\label{def:rel-cand-diag}
A \emph{relevant} candidate diagnosis of a symptom,
$\mathcal{S}=\tbeg  \Gamma_n, O^m_n \tend$, is a candidate diagnosis,
$\langle E,\Delta \rangle$, of $\mathcal{S}$ such that all actions in $E$
are relevant to the observations in $O^m_n$.
\end{definition}

\st
In the theorems that follow, we will implicitly consider only well-defined
system descriptions.
We will also write $\langle s_0, \alpha, s_1 \rangle \in T$,
to indicate that transition diagram $T$ contains a path from state $s_0$ to state $s_1$
whose actions are labeled by sequence of compound actions $\alpha$.
%
\begin{lemma}\label{th:rel-lemma-1}
Let $O$ be a collection of fluent literals, $s_0$, $s_0'$ be states such that
$s_0 \eqO{O} s_0'$, and $\alpha$ be a sequence of actions s.t. every action
of $\alpha$ is relevant to $O$.
If $\langle s_0, \alpha, s_1 \rangle \in T$, then there exists $s_1'$ such that
\begin{enumerate}
\item
$\langle s_0', \alpha, s_1' \rangle \in T$;
\item
$s_1' \eqO{O} s_1$.
\end{enumerate}
\end{lemma}
\begin{lemma}\label{th:rel-lemma-2}
Let $O$ be a collection of fluent literals, and $e, a_1, \ldots, a_k$ be elementary
actions. If $e$ is not relevant to $O$ 
and $\langle s_0, \{ e, a_1, \ldots, a_k \}, s_1 \rangle \in T$,
then there exists $s_1'$ such that
\begin{enumerate}
\item
$\langle s_0, \{ a_1, \ldots, a_k \}, s_1' \rangle \in T$;
\item
$s_1 \eqO{O} s_1'$.
\end{enumerate}
\end{lemma}
\begin{theorem}\label{th:rel-1}
Let $SD$ be a system description and $O$ be a collection of fluent literals.
For every path $\langle s_0, \alpha, s_f \rangle$ from $\mathcal{T}(SD)$ and
$s_0' \eqO{O} s_0$ there is a path $\langle s_0', \alpha', s_f' \rangle$
such that:
\begin{equation}\label{thesis:rel-1:1}
\alpha \eqO{O} \alpha'\mbox{, }|\alpha'|_O=0\mbox{;}
\end{equation}
\begin{equation}\label{thesis:rel-1:2}
s_f' \eqO{O} s_f\mbox{.}
\end{equation}
%
%
\Begproof
By induction on the rank of $\alpha$ w.r.t. $O$, $|\alpha|_O$. In the rest of this proof,
we will use $T$ to denote $\mathcal{T}(SD)$.

\st
Base case: $|\alpha|_O=0$. By Lemma \ref{th:rel-lemma-1}, there exists a path
$\langle s_0', \alpha, s_f' \rangle \in T$ such that $s_f' \eqO{O} s_f$.

\st
Inductive step: since $|\alpha|_O>0$, at least one elementary action irrelevant to
$O$ occurs in $\alpha$. Hence, there exist states $s_k$, $s_{k+1}$, elementary actions
$e_1, \ldots, e_m$, and (possibly empty) sequences of actions
$\alpha_1$, $\alpha_2$ such that:
\begin{enumerate}
\item\label{cond:th:rel-1}
all actions in $\alpha_1$ are relevant to $O$, and $\langle s_0, \alpha_1, s_k \rangle \in T$;
\item\label{cond:th:rel-2}
$e_1$ is not relevant to $O$;
\item\label{cond:th:rel-3}
$\langle s_k, \{ e_1, \ldots, e_m \}, s_{k+1} \rangle \in T$;
\item\label{cond:th:rel-4}
$\langle s_{k+1}, \alpha_2, s_f \rangle \in T$.
\end{enumerate}
First we show that there exists $s_{k+1}'$ such that
\begin{equation}\label{eq:th:rel-1:1}
\langle s_0', \alpha_1 \{ e_2, \ldots, e_m\}, s_{k+1}' \rangle \in T\mbox{.}
\end{equation}
and $s_{k+1}' \eqO{O} s_{k+1}$.

\st
Let $s_0'$ be a state such that $s_0' \eqO{O} s_0$. From condition (\ref{cond:th:rel-1})
and the inductive hypothesis, it follows that there exists $s_k' \eqO{O} s_k$ such that 
\begin{equation}\label{eq:th:rel-1:2}
\langle s_0', \alpha_1, s_k' \rangle \in T\mbox{.}
\end{equation}
Next, we notice that, by conditions (\ref{cond:th:rel-2}), (\ref{cond:th:rel-3}) and
Lemma \ref{th:rel-lemma-2},
there exists $s_{k+1}''$ such that $s_{k+1}'' \eqO{O} s_{k+1}$
and $\langle s_k, \{ e_2, \ldots, e_m \}, s_{k+1}'' \rangle \in T$.
By inductive hypothesis, there exists $s_{k+1}'$ such that $s_{k+1}' \eqO{O} s_{k+1}''$
and 
\begin{equation}\label{eq:th:rel-1:3}
\langle s_k', \{ e_2, \ldots, e_m \}, s_{k+1}' \rangle \in T\mbox{.}
\end{equation}
Since relation ``$\eqO{O}$'' is transitive, 
\begin{equation}\label{eq:th:rel-1:3a}
s_{k+1} \eqO{O} s_{k+1}'\mbox{.}
\end{equation}
Hence, (\ref{eq:th:rel-1:1}) is proven.

\st
Notice that, by construction, $|\alpha_2|_O < |\alpha|_O$.
By condition (\ref{cond:th:rel-4}), equation (\ref{eq:th:rel-1:3a}), and the inductive hypothesis,
we obtain that there exist $s_f''$ and $\alpha_2'$ such that $s_f' \eqO{O} s_f$,
$\alpha_2 \eqO{O} \alpha_2'$, $|\alpha_2'|_O=0$, and
\begin{equation}\label{eq:th:rel-1:4}
\langle s_{k+1}', \alpha_2', s_f'' \rangle \in T\mbox{.}
\end{equation}

\st
Let $\gamma$ be the sequence consisting of $\alpha_1$, $\{e_2, \ldots, e_m\}$
and $\alpha_2'$. From (\ref{eq:th:rel-1:2})--(\ref{eq:th:rel-1:4}), it follows that
\begin{equation}\label{eq:th:rel-1:5}
\langle s_o', \gamma, s_f'' \rangle \in T\mbox{.}
\end{equation}
By the inductive hypothesis there exists a path $\langle s_0', \alpha', s_f' \rangle \in T$
such that $\alpha' \eqO{O} \gamma$ and $s_f' \eqO{O} s_f''$. (\ref{thesis:rel-1:1})
and (\ref{thesis:rel-1:2}) follow immediately from the transitivity of relation ``$\eqO{O}$''.
\Endproof
\end{theorem}
The following theorem shows that, if only relevant candidate diagnoses are computed,
no useful prediction about the system's malfunctioning is missed.
\begin{theorem}\label{th:rel-1-1}
Let $\langle \Sigma, T, W \rangle$ be a diagnostic domain, $SD$ be a system
description of $T$, and $\mathcal{S}=\tbeg  \Gamma_n, O \tend$ be a symptom of the
system's malfunctioning. For every candidate diagnosis $D=\langle E, \Delta \rangle$
of $\mathcal{S}$ there exists a relevant candidate diagnosis
$D_r = \langle E_r, \Delta_r \rangle$ such that $E_r = red_{O}(E)$.

\Begproof
First, let us prove that $D_r$ is a candidate diagnosis.
By definition of candidate diagnosis (Definition \ref{d1}), $E$ describes one or more paths
in the transition diagram, whose final state, $s_f$, is consistent with $O$.
By Theorem \ref{th:rel-1}, $E_r$ describes paths whose final state, $s_f'$, is
equivalent to $s_f$ w.r.t. $O$. Hence, $s_f'$ is consistent with $O$, as well,
and therefore $D_r$ is a candidate diagnosis.

\st
The fact that $D_r$ is a relevant candidate diagnosis follows directly from
Definition \ref{def:rel-cand-diag}.
\Endproof
\end{theorem}
The next theorem
proves that diagnostic module $D_1(\mathcal{S})$ generates all relevant
candidate diagnoses of $\mathcal{S}$.
\begin{theorem}\label{th:rel-2}
Let $\langle \Sigma, T, W \rangle$ be a diagnostic domain, and $SD$ be a system
description of $T$. For every symptom of the system's malfunctioning,
$\mathcal{S}=\tbeg  \Gamma_n, O^m_n \tend$, diagnostic module $D_1(\mathcal{S})$
computes all relevant candidate diagnoses of $\mathcal{S}$.

\Begproof
(Sketch.) By Theorem \ref{theorem:candidate_diag}, $D_0(\mathcal{S})$ computes all
candidate diagnoses of $\mathcal{S}$. $D_1(\mathcal{S})$ essentially consists in the addition
of a constraint. This constraint makes the module reject all candidate diagnoses
which are not relevant to the observations in $O^m_n$.
Hence, all candidate diagnoses returned by $D_1(\mathcal{S})$ are relevant
to $\mathcal{S}$.
\Endproof
%
\end{theorem}

\subsection{Properties of $Find\_Diag$}
The properties of $Find\_Diag$ are described by Theorem
\ref{theorem:find_diag}. In order to prove the theorem, we prove separately the
termination of the algorithm and its correctness. 

\begin{lemma}\label{lemma:term-find_diag}
Let $\langle \Sigma, T, W \rangle$ be a diagnostic domain, $SD$ be a
system description of $T$, and $\mathcal{S}=\tbeg  \Gamma_n, O_n \tend$
be a symptom of the system's malfunctioning. Then, $Find\_Diag(\mathcal{S})$
terminates.
\if T\extPaper
\Begproof
Recall that $\Sigma=\langle C, F, A \rangle$. Let us prove by induction
that the following is an invariant of the repeat $\ldots$ until loop:

\st
``at the beginning of the $i^{th}$ iteration of the loop, there is a set $M
\subseteq C$ such that
\begin{equation}\label{eq:term-find_diag-invariant}
\begin{array}{l}
(a) \qquad |M| = i-1 \\
(b) \qquad \forall c \in M \,\, obs(\neg ab(c),m) \in O \\
\end{array}
\end{equation}
Base case: $i=1$. Trivially satisfied.

\st
Inductive step. Assume that (\ref{eq:term-find_diag-invariant}) holds for $i$
and prove that it holds for $i+1$. Let $M=\{c_1, \ldots, c_{i-1}\}$ be the set
satisfying (\ref{eq:term-find_diag-invariant}) for $i$. Since the loop was not
exited after the last iteration, we can conclude that, during
iteration $m$, a component $c$ was found, such that $observe(m,ab(c)) \not=
ab(c)$ and $c$ was hypothesized faulty in the candidate diagnosis computed by
$Candidate\_Diag$. Under these conditions, the \emph{else} branch inside the
while loop adds $obs(\neg ab(c),m)$ to $O$. Notice that, by definition of
candidate diagnosis, this literal did not previously belong to $O$, which
implies $c \not\in M$. Then $M \cup \{c\}$ satisfies
(\ref{eq:term-find_diag-invariant}) for $i+1$. 

\st
To complete the proof we show that the algorithm performs at most $|C|+1$
iterations of the repeat $\ldots$ until loop. Suppose it is in the beginning of
iteration $|C|+1$. From the invariant, we obtain that
\begin{equation}
\forall c \in C \,\, obs(\neg ab(c),m) \in O.
\end{equation}
Definition (\ref{d1}) implies that $\langle
\Gamma_n, O \rangle$ has no candidate diagnosis. The next call to
$Candidate\_Diag$ returns $\langle \emptyset, \emptyset \rangle$ and the
algorithm terminates immediately.
\Endproof
\fi
\end{lemma}

\begin{lemma}\label{lemma:corr-find_diag}
Let $SD$ be defined as above, $\mathcal{S}_0=\tbeg 
\Gamma_n, O^m_n \tend$ be a symptom of the system's malfunctioning, and
$\tbeg E, \Delta \tend= Find\_Diag(\mathcal{S})$, where the value of variable
$\mathcal{S}$ is set to $\mathcal{S}_0$. If $\Delta \not= \emptyset$, then
\[
\tbeg E, \Delta \tend \mbox{ is a diagnosis of } \mathcal{S}_0;
\]
otherwise, $\mathcal{S}_0$ has no diagnosis.

\Begproof
Let us show that, if $\Delta=\emptyset$, $\mathcal{S}_0$ has no
diagnosis. By definition of candidate diagnosis and Theorem
\ref{theorem:candidate_diag}, $Candidate\_Diag$ returns $\Delta=\emptyset$ only
if $\mathcal{S}_0$ has no diagnosis. The proof is completed by observing that,
if $Candidate\_Diag$ returns $\Delta=\emptyset$, the function terminates
immediately, and returns $\langle E, \Delta \rangle$.

\st
Let us now assume that $\Delta\not=\emptyset$. We have to show that
\begin{enumerate}
\item\label{th:corr-find_diag-1}
$\tbeg E, \Delta \tend$ is a candidate diagnosis of $\mathcal{S}_0$, and
\item\label{th:corr-find_diag-2}
all the components in $\Delta$ are faulty.
\end{enumerate}
Let $O^i$ ($i \ge 1$) denote the value of variable $O$ at the beginning of
the $i^{th}$ iteration of the $repeat \ldots until$ loop of $Find\_Diag$. Notice
that $O^1$ is equal to the initial value of $O_n$. By Lemma
\ref{lemma:term-find_diag}, the algorithm terminates. Let $u$ denote the index
of the last iteration of the $repeat \ldots until$ loop.

\st
Since $\Delta \not= \emptyset$, by Theorem \ref{theorem:candidate_diag} and
by the fact that $\langle E,\Delta \rangle$ is not updated by the while loop,
\[
\tbeg E, \Delta \tend \mbox{ is a candidate diagnosis of } \tbeg \Gamma_n,
O^u \tend.
\]
We want to show that this implies statement \ref{th:corr-find_diag-1}.

\st
By the definition of candidate diagnosis,
\begin{equation}\label{eq:corr-find_diag-1}
\begin{array}{c}
\mbox{there exists a model, $M$, of } \Gamma_n \cup O^u \cup E \mbox{ such
that} \\
\Delta=\{ c \,\,|\,\, M \entails h(ab(c),m) \}\mbox{.}
\end{array}
\end{equation}
Let $M$ denote one such model. From Corollary \ref{corollary:main_incomp},
(\ref{eq:corr-find_diag-1}) holds iff
\begin{equation}\label{eq:corr-find_diag-2}
\begin{array}{c}
\mbox{there exists an answer set, $\mathcal{AS}$, of }\\ P=\alpha(SD,
\Gamma_n \cup O^u \cup
E) \cup R \mbox{ such that } \\
\Delta=\{ c \,\,|\,\, h(ab(c),m) \in \mathcal{AS} \}\mbox{.}
\end{array}
\end{equation}
Let $A^u$ denote one such answer set.

\st
Since $\mathcal{S}_0$ is a symptom, $n>0$. Notice that $O'=O^u \setminus
O^1$ is a set of observations made at time $n$. Let $C$ denote the set of
constraints of $P$ of the form
\begin{equation}
\begin{array}{rlcl}
    &       & \hif  & obs(l,t), \\
    &       &   & \lpnot h(l,t)\mbox{.}
\end{array}
\end{equation}
where $obs(l,t) \in O'$ -- these constraints correspond to rule (10) of
$\Pi$ (see Section \ref{sec:computing})). Let also $Q$ denote $P \setminus C$.

\st
By the properties of the answer set semantics, (\ref{eq:corr-find_diag-2})
holds iff
\begin{equation}\label{eq:corr-find_diag-3}
\begin{array}{c}
A^u \mbox{ is an answer set of } Q, A^u \mbox{ does not violate } C, \mbox{
and} \\
\Delta=\{ c \,\,|\,\, h(ab(c),m) \in A^u \}\mbox{.}
\end{array}
\end{equation}
%
Notice that $O'$ is a splitting set for $Q$, and $bottom_{O'}(Q)=O'$. Since no
literal of $O'$ occurs in $top_{O'}(Q)$, $e_{O'}(Q,O')=Q \setminus O'$. Let $R$
denote $Q \setminus O'$. By the Splitting Set Theorem, $A^u$ is an answer set
of $Q$ iff $A^u \setminus O'$ is an answer set of $R$.

\st
Let $A^1$ denote $A^u \setminus O'$. Observe that the literals of $O'$
occur, within $P$, only in the constraints of $C$, and that they never occur
under negation as failure. Therefore, if $A^u$ does not violate $C$, then $A^1$
does not violate $C$, either. Hence, (\ref{eq:corr-find_diag-3}) implies that
\begin{equation}\label{eq:corr-find_diag-4}
\begin{array}{c}
A^1 \mbox{ is an answer set of } R, A^1 \mbox{ does not violate } C, \mbox{
and} \\
\Delta=\{ c \,\,|\,\, h(ab(c),m) \in A^1 \}\mbox{.}
\end{array}
\end{equation}
By the properties of the answer set semantics, (\ref{eq:corr-find_diag-4})
holds iff
\begin{equation}\label{eq:corr-find_diag-5}
\begin{array}{c}
A^1 \mbox{ is an answer set of } R \cup C, \mbox{ and} \\
\Delta=\{ c \,\,|\,\, h(ab(c),m) \in A^1 \}\mbox{.}
\end{array}
\end{equation}
Since $R=Q \setminus O'$,
\[
R \cup C = Q \setminus O' \cup C = P \setminus O' = \alpha(SD,\Gamma_n
\cup O^1 \cup E) \cup R\mbox{.}
\]
Hence (\ref{eq:corr-find_diag-5}) can be rewritten as:
\begin{equation}\label{eq:corr-find_diag-6}
\begin{array}{c}
A^1 \mbox{ is an answer set of } \alpha(SD,\Gamma_n \cup O^1 \cup E) \cup
R, \mbox{ and}\\
\Delta=\{ c \,\,|\,\, h(ab(c),m) \in A^1 \}\mbox{.}
\end{array}
\end{equation}

\st
From Corollary \ref{corollary:main_incomp}, (\ref{eq:corr-find_diag-6}) holds
iff 
\begin{equation}\label{eq:corr-find_diag-7}
\begin{array}{c}
\mbox{there exists a model, $M$, of } \Gamma_n \cup O^1 \cup E \mbox{ such
that}
\\
\Delta=\{ c \,\,|\,\, M \entails h(ab(c),n-1) \}\mbox{.}
\end{array}
\end{equation}
By the definition of candidate diagnosis,
\[
\tbeg E, \Delta \tend \mbox{ is a candidate diagnosis of } \mathcal{S}_0\mbox{.}
\]

\st
To prove statement \ref{th:corr-find_diag-2}, notice that $\Delta$
is the result of the latest call to $Candidate\_Diag$. Since, by hypothesis,
$\Delta \not= \emptyset$, the value of variable $diag$ at the end of the final
iteration of the $repeat \ldots until$ loop must have been \emph{true}. In turn,
this implies that, in the same iteration, the while loop terminated with
$\Delta_0=\emptyset$ and $diag=true$. Therefore all the components in
$\Delta$ are faulty and, by definition of diagnosis,
\[
\tbeg E, \Delta \tend \mbox{ is a diagnosis of } \mathcal{S}_0\mbox{.}
\]
\Endproof
\end{lemma}

\st
\emph{Theorem \ref{theorem:find_diag}}\\
Let $\langle \Sigma, T, W \rangle$ be a diagnostic domain, $SD$ be a
system description of $T$, and $\mathcal{S}=\tbeg  \Gamma_n, O^m_n \tend$
be a symptom of the system's malfunctioning. Then,
\begin{enumerate}
\item\label{th:find_diag-1}
$Find\_Diag(\mathcal{S})$ terminates;
\item\label{th:find_diag-2}
Let $\tbeg E, \Delta \tend = Find\_Diag(\mathcal{S})$, where the value of
variable $\mathcal{S}$ is set to $\mathcal{S}_0$. If $\Delta \not= \emptyset$,
then
\[
\tbeg E, \Delta \tend \mbox{ is a diagnosis of } \mathcal{S}_0;
\]
otherwise, $\mathcal{S}_0$ has no diagnosis.
\end{enumerate}

\Begproof
Statement \ref{th:find_diag-1} is proven by applying Lemma
\ref{lemma:term-find_diag}. Statement \ref{th:find_diag-2} is proven by applying
Lemma \ref{lemma:corr-find_diag}.
\Endproof

\normalsize

\end{document}